\documentclass{article}

\usepackage{arxiv}

\usepackage[utf8]{inputenc} 
\usepackage[T1]{fontenc}    
\usepackage{hyperref}       
\usepackage{url}            
\usepackage{booktabs}       
\usepackage{amsfonts}       
\usepackage{nicefrac}       
\usepackage{microtype}      
\usepackage{lipsum}		
\usepackage{graphicx}
\usepackage{doi}
\usepackage{amsthm}
\usepackage{amsmath}
\theoremstyle{plain}
\newtheorem{theorem}{Theorem}
\newtheorem{lemma}[theorem]{Lemma}

\newtheorem{proposition}[theorem]{Proposition}

\theoremstyle{definition}

\theoremstyle{remark}
\newtheorem{remark}{Remark}

\title{Reinforcement Learning in Hyperbolic Spaces: Models and Experiments}

\author{Vladimir Ja\' cimovi\' c \\
	Faculty of Natural Sciences and Mathematics\\
	University of Montenegro\\ 
	Cetinjski put bb., 81000 Podgorica\\ 
	Montenegro\\
	\texttt{vladimirj@ucg.ac.me} \\
	\And
        Zinaid Kapi\' c\\
        Faculty of Engineering\\
        University of Rijeka\\
        Vukovarska 58, 51000 Rijeka\\
        Croatia\\
        \texttt{zinaid.kapic@unbi.ba} \\
        \And
	Aladin Crnki\' c \\
	Faculty of Technical Engineering\\
	University of Biha\' c\\
	Irfana Ljubijanki\' ca bb., 77000 Biha\' c\\
	Bosnia and Herzegovina\\
	\texttt{aladin.crnkic@unbi.ba} \\
}

\renewcommand{\shorttitle}{Reinforcement Learning in Hyperbolic Spaces: Models and Experiments}

\hypersetup{
pdftitle={Reinforcement Learning in Hyperbolic Spaces: Models and Experiments},
pdfsubject={-},
pdfauthor={Vladimir Ja\' cimovi\' c, Zinaid Kapi\' c, Aladin Crnki\' c},
pdfkeywords={Poincar\' e disc, black-box optimization, hyperbolic data, stochastic policy},
}

\begin{document}
\maketitle

\begin{abstract}
We examine five setups where an agent (or two agents) seeks to explore unknown environment without any prior information. Although seemingly very different, all of them can be formalized as Reinforcement Learning (RL) problems in hyperbolic spaces. More precisely, it is natural to endow the action spaces with the hyperbolic metric. We introduce statistical and dynamical models necessary for addressing problems of this kind and implement algorithms based on this framework. Throughout the paper we view RL through the lens of the black-box optimization.
\end{abstract}


\section{Introduction}\label{sec:1}

With the explosive growth of machine learning techniques and applications, new paradigms and models with transformative power are enriching the field. One of the most remarkable trends in recent years is the rapid rise of significance of Riemannian geometry and Lie group theory. The underlying cause is the rising complexity of the data, motivating more sophisticated approaches, thus leading to the wide recognition that a great deal of data sets exhibit an intrinsic curvature. In other words, many data sets are naturally represented or faithfully embedded into non-Euclidean spaces. One apparent example of this kind are rotational motions in robotics. Rotations in $n$-dimensional space constitute the Lie group and do not have the structure of vector space. However, significance of non-Euclidean data extends very far beyond this particular example. Slightly less apparent, but even more ubiquitous are data representations in hyperbolic geometry. It is widely accepted that any data set with some (possibly hidden) hierarchical structure is naturally embedded into Riemannian manifolds with constant negative curvature \cite{NK1,NK2,SSGR}. Recent advances in various non-Euclidean representations of the data motivated systematic approaches thus giving rise to the emerging field named Geometric Deep Learning \cite{BBCV}.

Although the whole field is at early stage, investigations in hyperbolic ML are quite numerous and some fascinating results have already been reported. Some of the pioneering efforts in hyperbolic ML are made in physics, where hyperbolic data are ubiquitous due to the hyperbolic geometry of spacetime. For instance, machine learning of actions of the Lorentz groups (isometries in higher-dimensional hyperbolic spaces) are highly significant in particle physics \cite{BAORMK}. However, the potential of hyperbolic ML has already been examined in many fields, including NLP \cite{TBG}, recommender systems \cite{CHWDDV}, brain research \cite{Baker,BSS}, computer vision \cite{Mettes} and knowledge extraction \cite{Chami}. Furthermore, the novel architecture named {\it hyperbolic neural networks} has been proposed for problems of supervised hyperbolic ML \cite{GBH}.

Meanwhile, RL in hyperbolic spaces is at the very beginning. Recent preprint \cite{CCBH} provides arguments (supported by experiments) that most of RL problems can be viewed through the lens of hyperbolic ML. The underlying observation is that Markov decision processes are hierarchical structures which are naturally embedded into hyperbolic spaces. Hence, the design of RL policies can be efficiently implemented through explorations in a hyperbolic space. In the present paper we emphasize that in many setups it is advantageous to endow the action space with the hyperbolic metric. We pose several illustrative problems and introduce the statistical-geometric framework necessary for the design of stochastic policies. The main goal of the present study is twofold: to illustrate the ubiquity of hyperbolic RL; and to present adequate statistical models and optimization techniques for dealing with such problems. From the principled point of view, our study fits into the paradigm of {\it geometry-informed ML}, postulating that models and architectures are, to a great extent, enforced by geometry of the data.

We adopt the point of view on RL as black-box optimization, as in \cite{SHCSS} (exploration of the unknown environment by one or several agents).

Our exposition is built around several problems that are deliberately chosen tractable and transparent (with the dimension of the search space up to 20). We formalize each problem, introduce appropriate models and implement the algorithms.

In the next Section we explain five problems. In Section \ref{sec:3} we introduce statistical and dynamic models for the design of efficient algorithms. Implementations and solutions are presented in Section \ref{sec:4}. Finally, Section \ref{sec:5} contains concluding remarks and brief outlook on further extensions and potential applications.

\section{Problems}\label{sec:2}

In this Section we present five problems.

\subsection{Single-agent hyperbolic RL problems}

\subsubsection{The frog}\label{The_Frog_problem}
\begin{figure*}[h] \label{Frog1}
\centering
  \begin{tabular}{@{}c@{}}
    \includegraphics[width=0.75\textwidth]{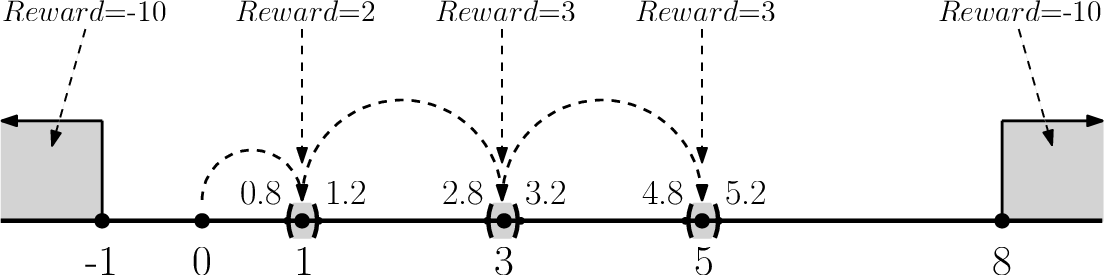}\\
  \end{tabular}
  \caption{\label{fig1}Illustration of the frog problem.}
\end{figure*}

Consider the game explained in the following lines. A player starts at zero and can make jumps of arbitrary lengths along the real line. They do not posses any prior information about the environment. Each jump costs one coin. There are three rewards worth 2,3 and 3 coins placed around points 1,3 and 5, respectively (see Figure \ref{fig1}). If the player jumps to a number which is less than $-1$ or greater than $8$, they incur a large penalty of ten coins.

The optimal solution is obvious: exactly three jumps of lengths 1, 2, and 2, resulting in the total reward 2+3+3-1-1-1=5, see Figure \ref{fig1}. Notice, that the player can collect rewards in a different order, hence jumps of the lengths 3,-2 and 4 also result in optimal reward.

One might find it surprising that we treat this game as a problem of hyperbolic RL. However, it makes a lot of sense to assume that the player explores their environment by random jumps, sampled from one-variate normal (Gaussian) distribution ${\cal N}(a,\sigma^2)$. Then the policies are encoded by several normal distributions, that is - by their parameters $m_i,\sigma_i^2$ for $i=1,\dots,p$. In fact, only 3 jumps are needed, but since the player does not have any prior information, they must learn an optimal number of jumps as well.

Now, a reader familiar with information geometry may recall that the Fisher information distance endows the family of one-variate normal distributions with the negative curvature. More precisely, by introducing the Fisher information metric on this family we turn it into statistical manifold that is isomorphic to the unit hyperbolic disc. We refer to \cite{CSS} for detailed explanation of this fact and the corresponding formulae. Hence, our problem can be formalized as learning an optimal configuration of $p$ points in the Poincar\' e disc.

\subsubsection{Embedding multi-layer graph into hyperbolic disc}

Hyperbolic ML is to a great extent motivated by representations of complex networks in hyperbolic spaces. These representations are based on the notion of {\it geometric graph}. The general idea is that one can map nodes of a graph to points in some space in such a way that the distance between any two points is less than $\epsilon>0$ if and only if there exists a link between the corresponding nodes \cite{KPKVB,BFKL} (where $\epsilon$ is a predefined threshold). An embedding which satisfies the above criterion is named {\it proper embedding} of a network. Notice that a proper embedding allows for an exact retrieval of the graph, since the whole information is preserved in configuration of points.

\begin{figure*}[h] \label{multi-tree}
\centering
  \begin{tabular}{@{}c@{}}
    \includegraphics[width=0.8\textwidth]{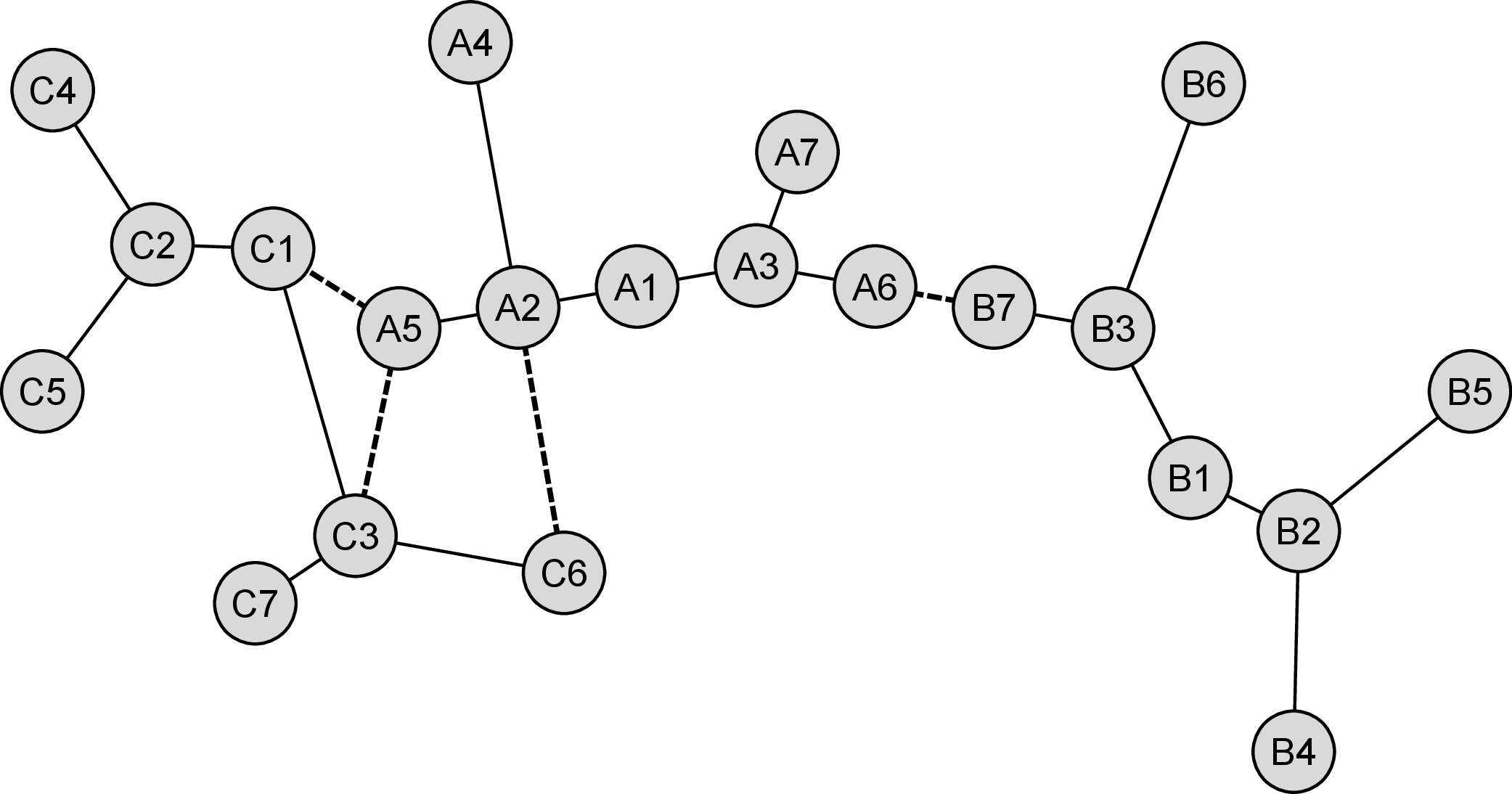}\\
  \end{tabular}
  \caption{\label{fig2}Multi-layer graph with three layers.}
\end{figure*}

Here, we consider embeddings of multi-layer graphs. Suppose that three trees denoted by A,B and C are properly embedded into hyperbolic disc. Then several links are established between nodes belonging to different trees (layers). In such a way we obtain a {\it multi-layer tree}, where new links between layers are represented by dashed lines (see Figure \ref{fig2}). We now aim to adjust our embeddings in such a way to take new intra-layer links into account without loosing prior information encoded in embeddings of the three trees.

Our goal now is to transform the set of points which corresponds to each tree by an isometry in the hyperbolic unit disc, in such a way to get those pairs of nodes connected by intra-layer links sufficiently close. By applying isometries we can be sure that previous information (regarding each separate layer) will be preserved. Underline that we are looking for a different isometry for each layer, since applying the same isometry to all layers does not change anything. In total, we aim to learn three coupled isometries of the hyperbolic unit disc. Notice that we can set one isometry at identity, thus leaving one layer intact. Then it remains to transform two layers only, by learning two coupled isometries of the unit disc. In order to address this problem, we will introduce appropriate models in Section \ref{sec:3}.

\subsubsection{Exploring directional labyrinth}

Labyrinths and mazes are among the favorite toys of RL. RL approaches are often validated on explorations of labyrinths. The most common are rectangular labyrinths, where the player chooses one of four directions (up, down, left, right) at each step. Here, we consider the labyrinth of another kind, that we name {\it directional labyrinth}. At each time-step the player can choose an orientation in space and moves in that direction. Hence, this is the labyrinth with continuous non-Euclidean search space. We are not aware of any previous study on labyrinths of this kind.

\begin{figure*}[h] \label{labyrinth}
\centering
  \begin{tabular}{@{}c@{}}
    \includegraphics[width=0.5\textwidth]{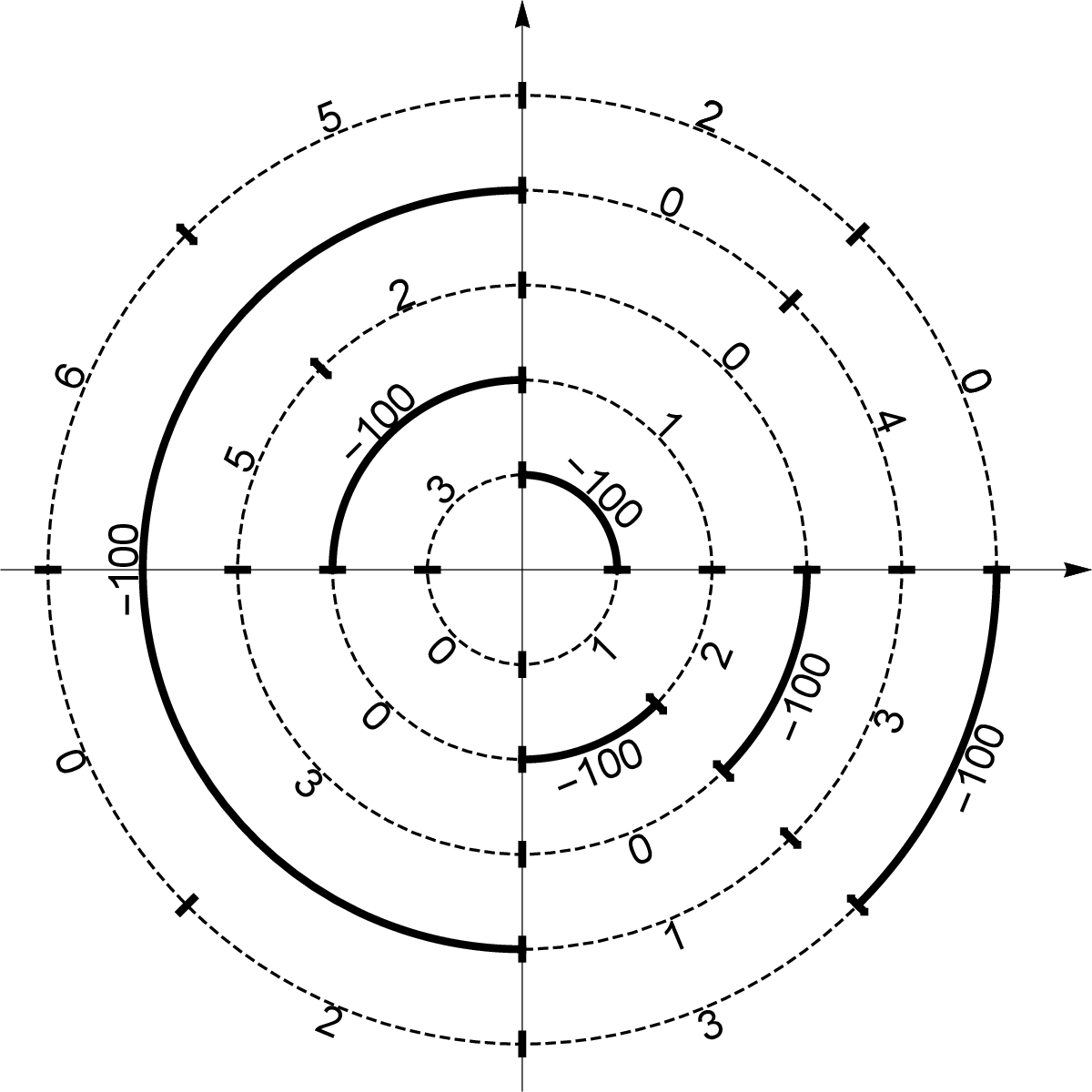}\\
  \end{tabular}
  \caption{\label{fig3}Reward setting in the directional labyrinth.}
\end{figure*}

The labyrinth with five concentric circles of radii 1,2,3,4,5 and the reward system is displayed in Figure \ref{fig3}. Consider the game with the following rules:

\begin{enumerate}

\item Player starts from the center of concentric circles.

\item At each step, the player chooses a direction (i.e. an angle) and makes the step of length one in that direction.

\item The player will make exactly 10 steps.

\item The player receives its reward by stepping on an arch of a circle where the rewards are written in Figure \ref{fig3}.

\item Each reward can be taken only once.

\item Along with rewards there are also penalties of $-100$. Intersecting an arc with the negative reward $-100$ costs $100$ coins. In essence these are prohibitive penalties turning these arcs into barriers.

\item The exceptions are intersections of circles on arcs with the negative reward of $-100$, which are explained in the previous point.

\end{enumerate}

Obviously, the problem consists in finding a sequence of ten angles. The search space is $10$-dimensional torus. In order to encode stochastic policies a certain family of probability distributions on the circle is required.

\begin{figure*}[h] \label{Field}
\centering
  \begin{tabular}{@{}c@{}}
    \includegraphics[width=0.5\textwidth]{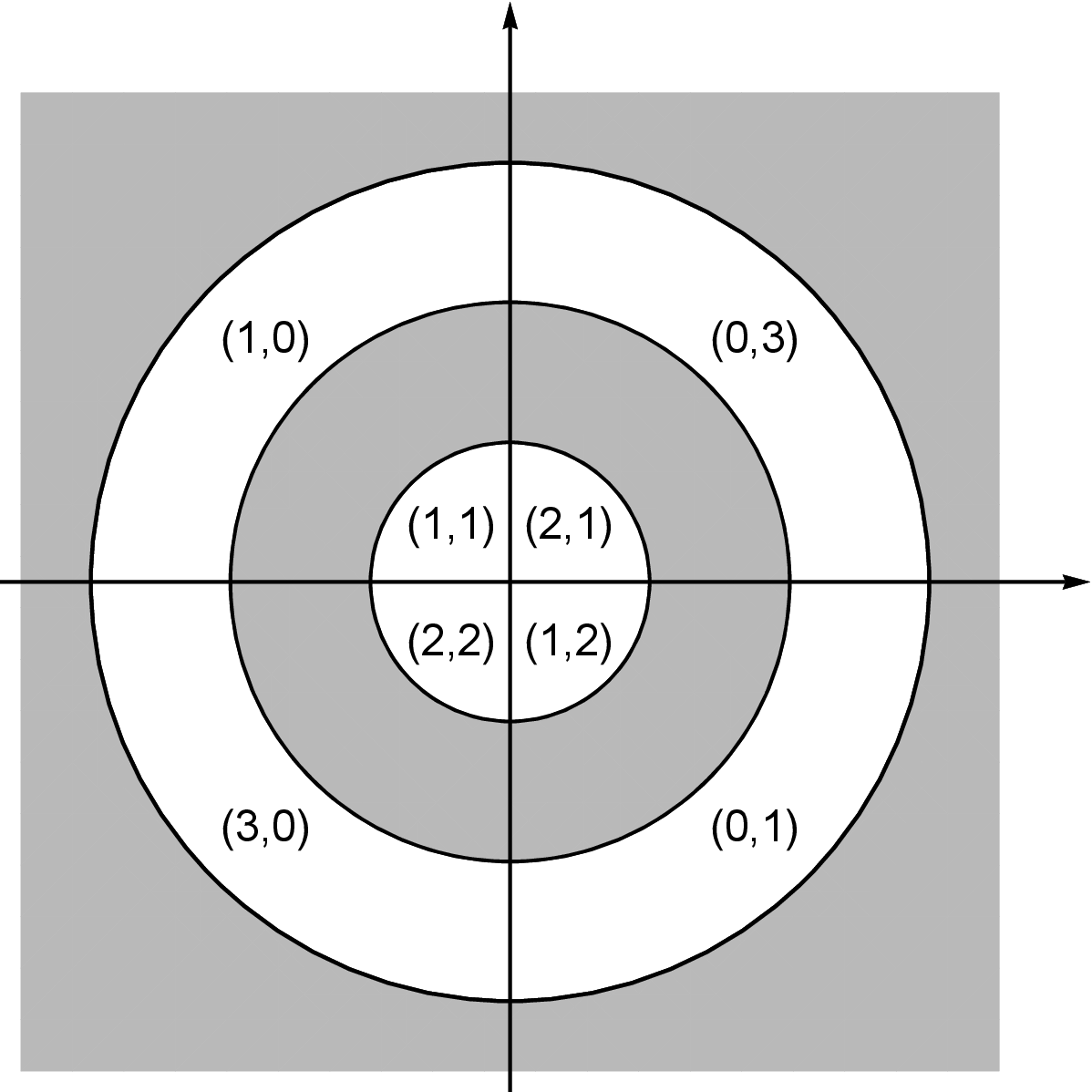}\\
  \end{tabular}
  \caption{\label{fig4}Reward setting of the problem \ref{two_agents_plane}.}
\end{figure*}

\subsection{Two-agents hyperbolic RL problems}

We proceed with two problems where two agents learn and adapt their strategies simultaneously.

\subsubsection{Two frogs}

Consider a modification of the frog problem described in \ref{The_Frog_problem}, by assuming that there are two frogs, and the game is extended by several additional rules:
\begin{enumerate}

\item The two players jump simultaneously.

\item There are two rewards at each spot, one for each player.

\item If the both players land at the same spot at the same time none of them receives the reward. The rewards remain intact in that situation.

\item The jumps shorter than 3 cost one coin, as before, but jumps longer than 3 cost 2 coins.

\end{enumerate}

In such a setup players tend to balance their strategies by maximizing their own rewards while avoiding conflicts with the other. Obviously, an optimal solution where both players receive the total reward of 5 is not feasible anymore. At least one player must miss some rewards or make jumps longer than 3, thus paying higher costs.

\subsubsection{Two agents exploring environment in the plane}
\label{two_agents_plane}

Consider the game of two players, determined by the following rules:

\begin{enumerate}

\item The two players select 3 numbers each: $a_1, a_2,a_3$, and $b_1,b_2, b_3$. This yields three vectors in the plane: $(a_1, b_1), (a_2, b_2)$ and $(a_3, b_3)$.

\item Denote by $\vec{p} = (a_1 + a_2 + a_3, b_1 + b_2 + b_3)$ the sum of these three vectors. If $\vec{p}$ falls on certain fields, the players receive rewards. The reward system is illustrated in Figure \ref{fig4}, where the two numbers represent rewards of the first and the second player respectively.

\item Figure \ref{fig4} shows the reward fields bordered by concentric circles of radii $1,2$ and $3$. There are no rewards in shaded areas of Figure \ref{fig4}. Thus, players receive rewards only if length of the vector $\vec{p}$ is between 0 and 1 or between 2 and 3.

\end{enumerate}

\section{Models}\label{sec:3}

In this Section we introduce some dynamical and statistical models which will be used in algorithms presented in Section \ref{sec:4}.

\subsection{Swarms}

We start with swarming dynamics in the unit disc. Swarms under consideration consist of individuals that we will refer to as "complex oscillators" (such a terminology is inspired by the Kuramoto model \cite{Kuramoto}). The position of each complex oscillator at a certain moment $t$ is represented by a point in the unit disc, that is - by a complex number $z_j = r_j e^{i \varphi_j}$ with $r_j = |z_j| <1$.
We use the term {\it Poincar\' e swarm} for such dynamical systems, because of relation with the Poincar\' e model of hyperbolic geometry in the unit disc.

The Poincar\' e swarm consisting of $N$ complex oscillators is governed by the following system of complex-valued ODE's:
\begin{equation}
\label{swarm}
\frac{d}{dt} z_j = i (f_j z_j^2 + \omega z_j + \bar f_j), \quad j=1,\dots,N,
\end{equation}
where
$$f_j = \frac{i}{2N} \sum \limits_{k=1}^N e^{i \beta_{jk}}K_{jk} \bar z_k.$$

Here, $\omega \in \mathbb{R}$ is generalized "frequency" of complex oscillators (we assume that all oscillators are identical, i.e. they all have the same intrinsic "frequency"). Furthermore, $K_{jk} \in \mathbb{R}$ and $\beta_{jk} \in [0,2 \pi)$ are respectively the coupling intensity and phase shift between complex oscillators $j$ and $k$. The couplings $K_{jk}$ can be negative or positive. In the present paper, we deal with symmetric interactions only: $K_{jk} = K_{kj}, \; \beta_{jk} = \beta_{kj} \; \forall j,k = 1,\dots,N$.

Dynamics \ref{swarm} preserve the unit disc, meaning that $|z_j(0)| < 1, \; \forall j=1,\dots,N$ implies that $|z_j(t)| < 1, \; \forall j=1,\dots,N$.

Now, consider the simplest Poincar\' e swarm with global coupling
\begin{equation}
\label{global_swarm}
\frac{d}{dt} z_j = i (f z_j^2 + \omega z_j + \bar f), \quad j=1,\dots,N,
\end{equation}
where
$$f = e^{i \beta} \frac{iK}{2N} \sum \limits_{k=1}^N \bar z_k.$$

The above system of ODE's describes the globally coupled swarm, where all couplings $K \in \mathbb{R}$ are of the same (negative or positive) intensity and all phase shifts $\beta$ are the same.

Furthermore, consider the group $M$ all conformal mappings (linear-fractional transformations) of the complex plane $\mathbb{C}$. Denote by $G$ the subgroup of $M$ consisting of all conformal mappings that preserve the unit disc $\mathbb{B}^2$. Finally, consider the group $G_+ \subset G$ of orientation preserving mappings. The group $G_+$ consists of transformations of the form:
\begin{equation}
\label{Mobius}
g(z) = e^{i \varphi} \frac{a-z}{1 - \bar a z}, \quad a \in \mathbb{B}^2, \; \varphi \in [0,2 \pi).
\end{equation}
Notice that the group $G_+$ is the Lie group of a real dimension three, parametrized by an angle $\varphi \in [0,2 \pi)$ and a point $a$ in the unit disc.

By adapting the result from \cite{MMS} we claim that the Poincar\' e swarm \ref{global_swarm} evolves by actions of one-paramteric group of conformal transformations of the form \ref{Mobius}. More precisely, the following assertion holds.

\begin{proposition} \label{evol_swarm}
Let $z_1(t),\dots,z_N(t)$ be trajectories satisfying the system \ref{global_swarm}. There exists a one-parametric family $g_t$ of transformations of the form \ref{Mobius}, such that
$$
z_j(t) = g_t(z_j(0)), \quad j=1,\dots,N, \; t>0.
$$
\end{proposition}
Recall that transformations of the form \ref{Mobius} are isometries of the unit disc with respect to hyperbolic metric \cite{Needham}. Hence, the above Proposition implies that the system \ref{global_swarm} preserves the distances and evolves by isometries of the hyperbolic disc $\mathbb{B}^2$.

The above Proposition also explains why Poincar\' e swarms provide an appropriate model for encoding isometries in the Poincar\' e disc. Moreover, swarms can be designed in such a way to encode actions of several coupled isometries. To that end, consider the swarm consisting of $p$ globally coupled sub-swarms:
\begin{equation}
\label{sub_swarms}
\frac{d}{dt} z_j^{(l)} = i (f (z_j^{(l)})^2 + \omega z_j^{(l)} + \bar f), \quad
j=1,\dots,N_l, \; l = 1,\dots,p,
\end{equation}
where
$$f = \frac{i}{2} \sum \limits_{m=1}^p e^{i \beta_{lm}}\frac{K_{lm}}{ N_m} \sum \limits_{k=1}^{N_m} \bar z_k^{(m)}.$$

Here, $N_1,\dots,N_p$ are numbers of complex oscillators in each sub-swarm and $N_1 + \cdots + N_p = N$ is total number of oscillators. $K_{lm} \in \mathbb{R}$ and $\beta_{lm} \in [0,2 \pi)$ are pairwise interactions between any pair of complex oscillators belonging to sub-swarms $l$ and $m$ respectively. Accordingly, $K_{ll}$ and $\beta_{ll}$ are interactions within the sub-swarm $l$, that is - coupling strength between two complex oscillators belonging to the same sub-swarm $l$.

\begin{proposition}\label{sub-swarm_evol}
Let $z_j^{(l)}, \; j=1,\dots,N_l, \; l=1,\dots,p$ be trajectories satisfying the system \ref{sub_swarms}. There exist $p$ one-parametric families $g_t^{(l)}$ of transformations of the form \ref{Mobius}, such that
$$
z_j^{(l)}(t) = g_t^{(l)}(z_j^{(l)}(0)), \quad j=1,\dots,N_l, \; l=1,\dots,p, \; t>0.
$$
\end{proposition}
Both propositions \ref{evol_swarm} and \ref{sub-swarm_evol} can be proven by slightly adapting the proof that the globally coupled ensemble of Kuramoto phase oscillators evolve by the actions of one-parametric families of conformal transformations on the unit circle. This proof (in fact, two different proofs) can be found in \cite{MMS}.

\subsubsection{Methods for training swarms}

As shown above, the Poincar\' e swarms encode isometries of the hyperbolic disc and thus provide an appropriate model for learning hyperbolic data.

In order to have the mathematical framework completed, we also need the methods for training the swarms. This is the question of learning the network of interactions which merits an extensive (both theoretical and experimental) study. Indeed, one could think of various methods including maximum likelihood estimation and score matching. In algorithms presented in the next Section, we use evolutionary optimization, more precisely the famous algorithm named Covariance Matrix Adaptation - Evolution Strategy (CMA-ES), see \cite{HO}. We take an advantage of the fact that couplings $K_{jm}$ in Poincar\' e swarms are real numbers. In addition, phase shifts $\beta_{jm}$ are angles, which can also be treated (with an insignificant loss of efficiency) as Euclidean variables. Hence, the space of parameters can be treated as Euclidean vector space thus providing an adequate framework for the CMA-ES algorithm. Therefore, by encoding hyperbolic isometries in swarming dynamics, we pass from optimization and learning in hyperbolic spaces to optimization in Euclidean spaces. In its turn, CMA-ES operates on Euclidean spaces and is the most successful stochastic search method in such a setup.

\subsection{Statistical models}

In order to design stochastic policies for addressing our problems we will employ three families of distributions on different domains. From the point of view of information geometry \cite{AJLS} they share one common property. Viewed as statistical manifolds all three of them are hyperbolic disc. This means that Fisher information distance between two distributions belonging to the same family coincides with the hyperbolic distance in the unit disc.

\subsubsection{Hyperbolic geometry of one-variate Gaussian distributions}

The best known probability distributions on the continuous space are normal (Gaussian) distributions on the real line. We denote this family by ${\cal N}(m,\sigma^2)$, where $m \in \mathbb{R}$ and $\sigma^2>0$ are mathematical expectation and the variance, respectively. We see that the family ${\cal N}(m,\sigma^2)$ is two-dimensional.

By introducing the Fisher information distance (induced by the Kullback-Leibler divergence) on ${\cal N}(m,\sigma^2)$ we find that this statistical manifold is isomorphic to the hyperbolic half-space and further (using the Cayley transformation) to the hyperbolic unit disc. For information-geometric point of view on this family we refer to \cite{CSS}. In the next Remark we briefly explain the mapping from hyperbolic disc to the manifold ${\cal N}(m,\sigma^2)$.

\begin{remark} \label{hyp_Gauss}
Let $\zeta$ be a complex number with $|\zeta|<1$, corresponding to a point in the unit disc $\mathbb{B}^2$. Transform $\zeta$ by the inverse Cayley transformation:
$$
\xi = i \frac{1-\zeta}{1+\zeta}.
$$
Then $\xi$ is a point in the upper half-plane. Let $m \in \mathbb{R}$ and $\sigma^2 > 0$ be numbers defined by the following relations:
$$
m = Re(\xi), \; \sigma^2 = \frac{1}{\sqrt{2}} Im(\xi).
$$
where notations $Re$ and $Im$ stand for the real and imaginary part of a complex number.
\end{remark}

In such a way, we constructed an invertible map: $\chi : \mathbb{B}^2 \to {\cal N}(m,\sigma^2)$. This correspondence between points in the hyperbolic disc and normal distributions will be used in solutions to some of the problems introduced in Section \ref{sec:2}.

\begin{remark}
Recall that multi-variate Gaussians ${\cal N}(a,\Sigma)$ are parametrized by the vector of mathematical expectation $a$ and positive-definite covariance matrix $\Sigma$. This family exhibits hyperbolic geometry (negative curvature) as well, but this statistical manifold is not isomorphic to a hyperbolic ball, as one might expect.
\end{remark}

\subsubsection{Wrapped Cauchy distributions on the circle}

One of the basic statistical models on the circle are so-called wrapped Cauchy distributions, introduced by Peter McCullagh in 1996 \cite{McCullagh}. These distributions are defined by densities of the form:
\begin{equation}
\label{wrapped_Cauchy}
p_{wC}(\varphi,r,\Phi) = \frac{1}{2 \pi} \frac{1-r^2}{1 + 2 r \cos (\varphi - \Phi) + r^2}, \quad \varphi \in \mathbb{S}^1.
\end{equation}
where $0 \leq r<1$ and $\Phi \in [0,2 \pi)$ are parameters of the distribution. In fact, we will parametrize densities \ref{wrapped_Cauchy} by one point in the disc $a = r e^{i \Phi}$.

We will denote the family defined by densities \ref{wrapped_Cauchy} by ${\cal WC}(a)$ with $a \in \mathbb{B}^2$. Wrapped Cauchy provides a simple example of a probability distribution on the Riemannian manifold. In general, probabilities on spheres and rotation groups are studied by the field of Directional Statistics \cite{MJ,PG-P}. One might expect wrapped Cauchy to be the most popular probability distributions on the circle, but there are also von Mises distributions which are more frequently employed in various problems of statistical modeling. Nevertheless, the family ${\cal WC}(a)$ has one conceptual advantage, substantiated in the following

\begin{lemma}
Wrapped Cauchy are obtained by the disc-preserving conformal mappings \ref{Mobius} of the uniform distribution on the unit circle $\mathbb{S}^1$.
\end{lemma}

It follows that the family ${\cal WC}(a)$ is invariant w.r. to actions of the group of conformal mappings that preserve the unit disc. Moreover, the group $G_+$ acts transitively on this family, meaning that for any two wrapped Cauchy distributions $wC(a_1)$ and $wC(a_2)$, there exists $g \in G_+$, such that $g_* wC(a_1) = wC(a_2)$.
(Here, $g_*$ denotes the pull-back measure, i.e. for a measure $\mu$ on the unit circle $g_* \mu(A) = \mu(g^{-1}(A))$ for any Borel set $A \subset \mathbb{S}^1$.)

The previous Lemma, along with Proposition 1 implies that uncertainties on the unit circle described by distributions \ref{wrapped_Cauchy} can be encoded (and, hence, learned by) in dynamics of Poincar\' e swarms.

Notice that the uniform distribution on $\mathbb{S}^1$ belongs to ${\cal WC}(a)$, obtained when setting $a=0$. Furthermore, delta distributions on $\mathbb{S}^1$ arise as limiting cases, when $r = |a| \to 1$.

By equipping the family ${\cal WC}(a)$ by Fisher information metric we obtain that this statistical manifold is isomorphic with the unit hyperbolic disc (see for instance \cite{AG}).

\subsubsection{Conformally natural distributions on the hyperbolic disc}

We will also use probability distributions on hyperbolic disc introduced in \cite{JM}. These distributions are defined by densities
\begin{equation}
\label{JacMark}
p_{CN}(\zeta) = \frac{s-1}{\pi} \left( \frac{(1 - |a|^2)(1 - |\zeta|^2)}{|1-\bar a \zeta|^2}\right)^s, \quad \zeta \in \mathbb{B}^2.
\end{equation}
Here, $a \in \mathbb{B}^2$ and $s>1$ are parameters of the distribution.

We denote this family by ${\cal CN}(a,s)$. It is invariant w. r. to actions of the group $G_+$. Moreover, for fixed $s=s_0$ this group acts transitively on the family ${\cal CN}(a,s_0)$:

\section{Formalizations and implementations}\label{sec:4}

In this Section we put the problems introduced in Section \ref{sec:2} in more rigorous mathematical framework, select an adequate model for each problem and present the solutions found by our algorithms.

\subsection{The frog}

We try to solve this problem with (no more than) five jumps. In order to learn that some jumps might be unnecessary, we do not penalize jumps shorter than $0.2$ (i.e. all jumps longer than $0.2$ cost 3 coins). In such a way, the problem is formalized as learning five normal distributions ${\cal N}(m_1,\sigma_1^2),\dots,{\cal N}(m_5,\sigma_5^2)$.

\noindent{\bf The algorithm}

\begin{enumerate}
\item[i)] Sample $5$ random points in hyperbolic disk $\mathbb{B}^2$ from the distribution \ref{JacMark} with $a=0$ and $s=2$.

\item[ii)] Map each of these points to parameters of normal distributions \\
$(m_1, \sigma_{1}^{2}),...,(m_5, \sigma_{5}^{2})$ as explained in Remark \ref{hyp_Gauss}.

\item[iii)] Sample $5$ random numbers $x_1,\dots,x_5$ from each of the five distributions obtained in step ii).

\item[iv)] Evaluate the reward for jumps of the lengths $x_1,\dots,x_5$. Repeat steps iii) and iv) 10 times, and calculate the average reward from ten evaluations.

\item[v)] Run the Poincar\' e swarm \ref{swarm} with five complex oscillators, with the initial conditions $z_1(0),\dots,z_5(0)$ sampled in step i). Initially couplings $K_{jk}$ and phase shifts $\beta_{jk}$ are sampled from $5$-variate standard normal distributions.

\item[vi)] Train the Poincar\' e swarm \ref{swarm} using the CMA-ES algorithm. For evaluations of the total reward, run the swarm \ref{swarm} on time interval $t \in [0,2]$ thus obtaining new $5$ points in $\mathbb(B)^2$.

\item[vii)] Repeat the steps ii)-iv) until the reward remains the same in several consecutive simulations.
\end{enumerate}

\begin{figure*}[h]
\centering
  \begin{tabular}{@{}ccc@{}}
    \includegraphics[width=0.32\textwidth]{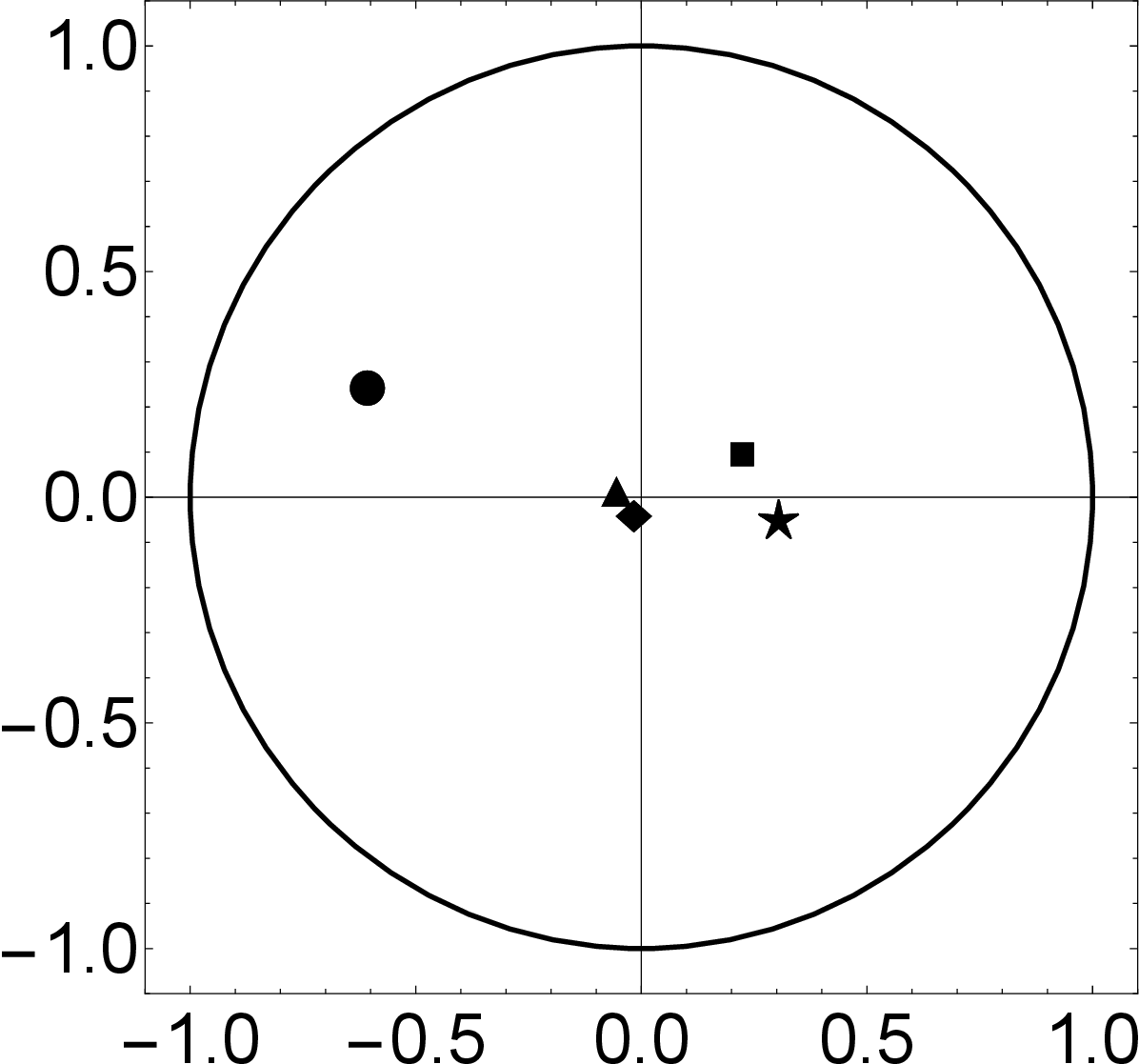}&\includegraphics[width=0.32\textwidth]{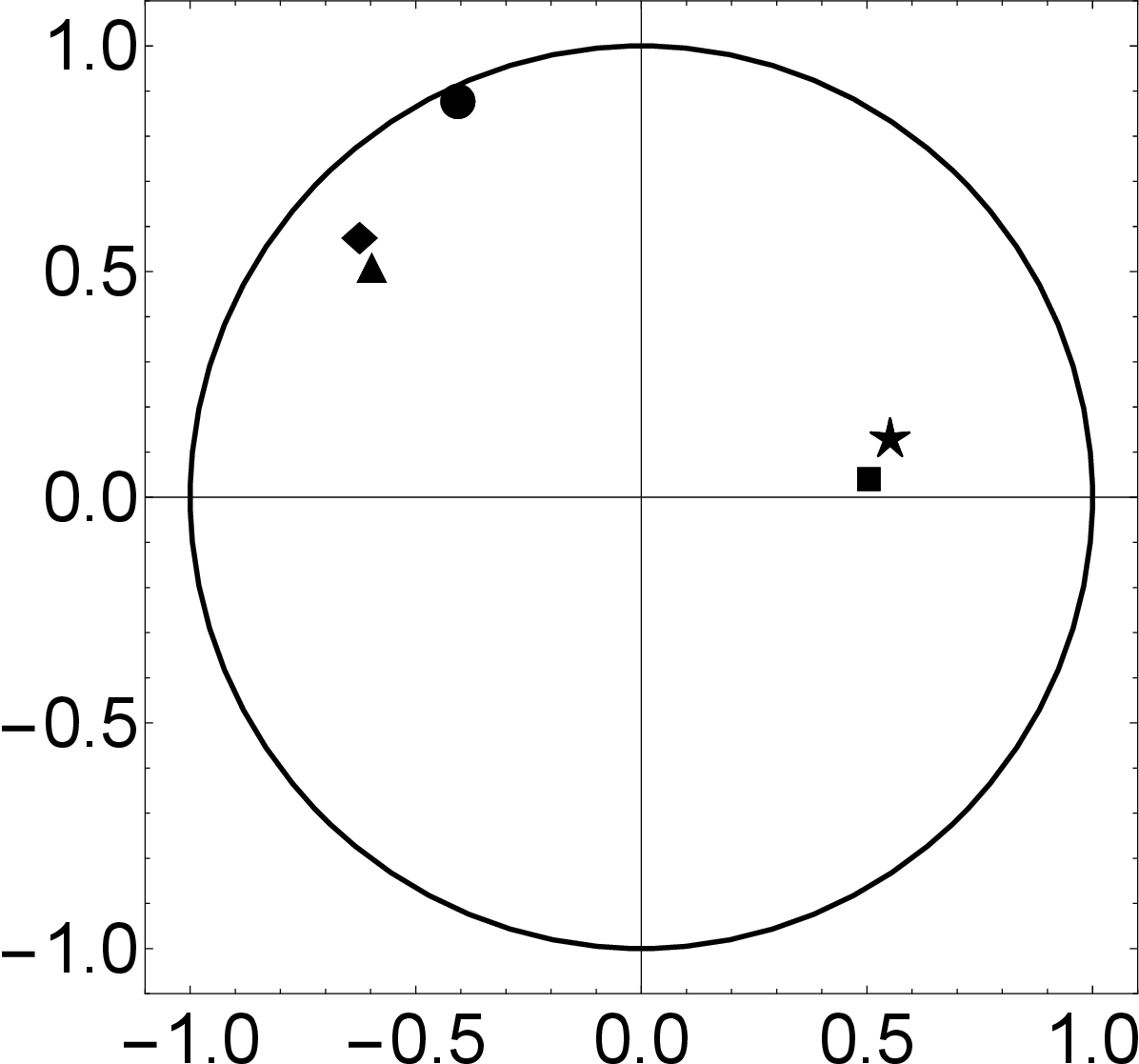}&\includegraphics[width=0.32\textwidth]{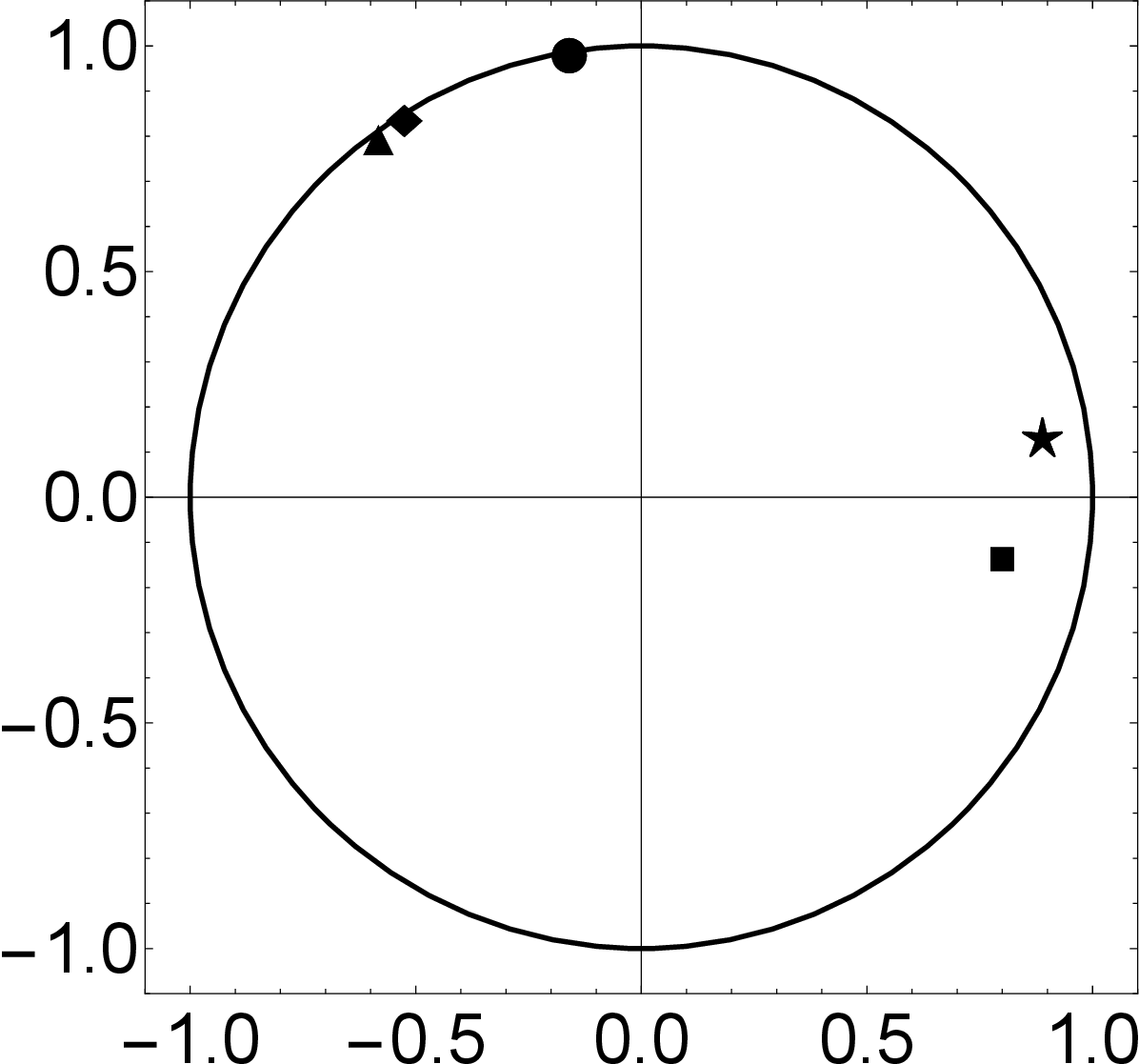}\\
    a)&b)&c)
  \end{tabular}
  \caption{\label{fig5}Position of points at the moments $T=0,1,2$ with corresponding distributions: a) ${\cal N}(2.31, 1.85)$, ${\cal N}(-0.07, 0.73)$, ${\cal N}(-0.06, 0.37)$, ${\cal N}(0.05, 0.79)$, ${\cal N}(0.12, 0.44)$; b) ${\cal N}(1.56, 0.03)$, ${\cal N}(2.42, 0.39)$, ${\cal N}(0.11, 0.20)$, ${\cal N}(2.41, 0.61)$, ${\cal N}(0.03, 0.23)$, and c) ${\cal N}(1.17, 0.00)$, ${\cal N}(1.80, 0.01)$, ${\cal N}(0.07, 0.04)$, ${\cal N}(1.96, 0.01)$, ${\cal N}(-0.08, 0.07)$. }
\end{figure*}

The final result are $5$ points in $\mathbb{B}^2$ that correspond to $5$ normal distributions (stochastic policies) that yield optimal jumps for the frog problem. Starting from fully stochastic policies, the algorithm gradually reduces the randomness, by decreasing the variances $\sigma_i^2$, see Figure \ref{fig5}. Such (almost) deterministic policies correspond to points placed near the circle, as in Figure \ref{fig5}c).

The algorithm always yields two points with mathematical expectation less than $0.2$, corresponding to two free jumps. Hence, the algorithm learns that an optimal reward must be taken in precisely three jumps.

\begin{figure*}[h]
\centering
  \begin{tabular}{@{}c@{}}
    \includegraphics[width=0.5\textwidth]{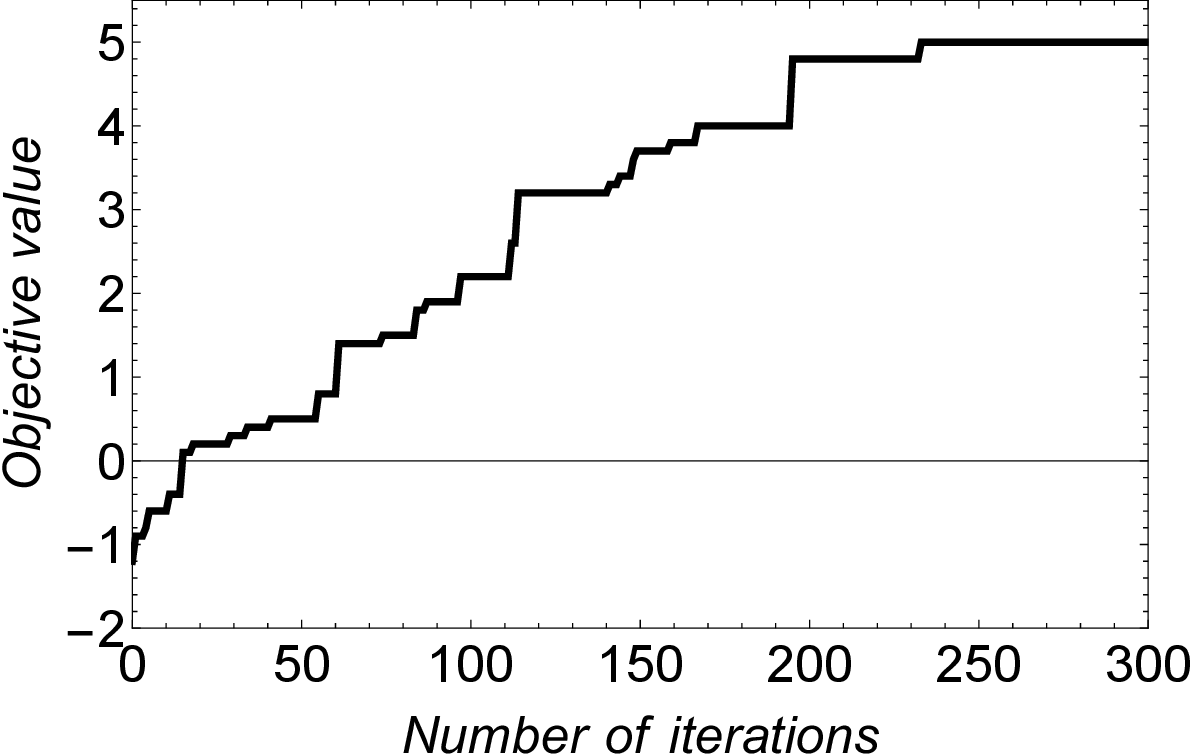}\\
  \end{tabular}
  \caption{\label{fig6} Evolution of the reward function of the frog problem.}
\end{figure*}

The evolution of the reward function is shown in Figure \ref{fig6}.

\subsection{Embedding multi-layer graph into hyperbolic disc}

This problem can be formalized as looking for two coupled isometries $g_1,g_2 \in G_+$ of the unit disc $\mathbb{B}^2$.

\begin{figure*}[h]
\centering
  \begin{tabular}{@{}ccc@{}}
    \includegraphics[width=0.32\textwidth]{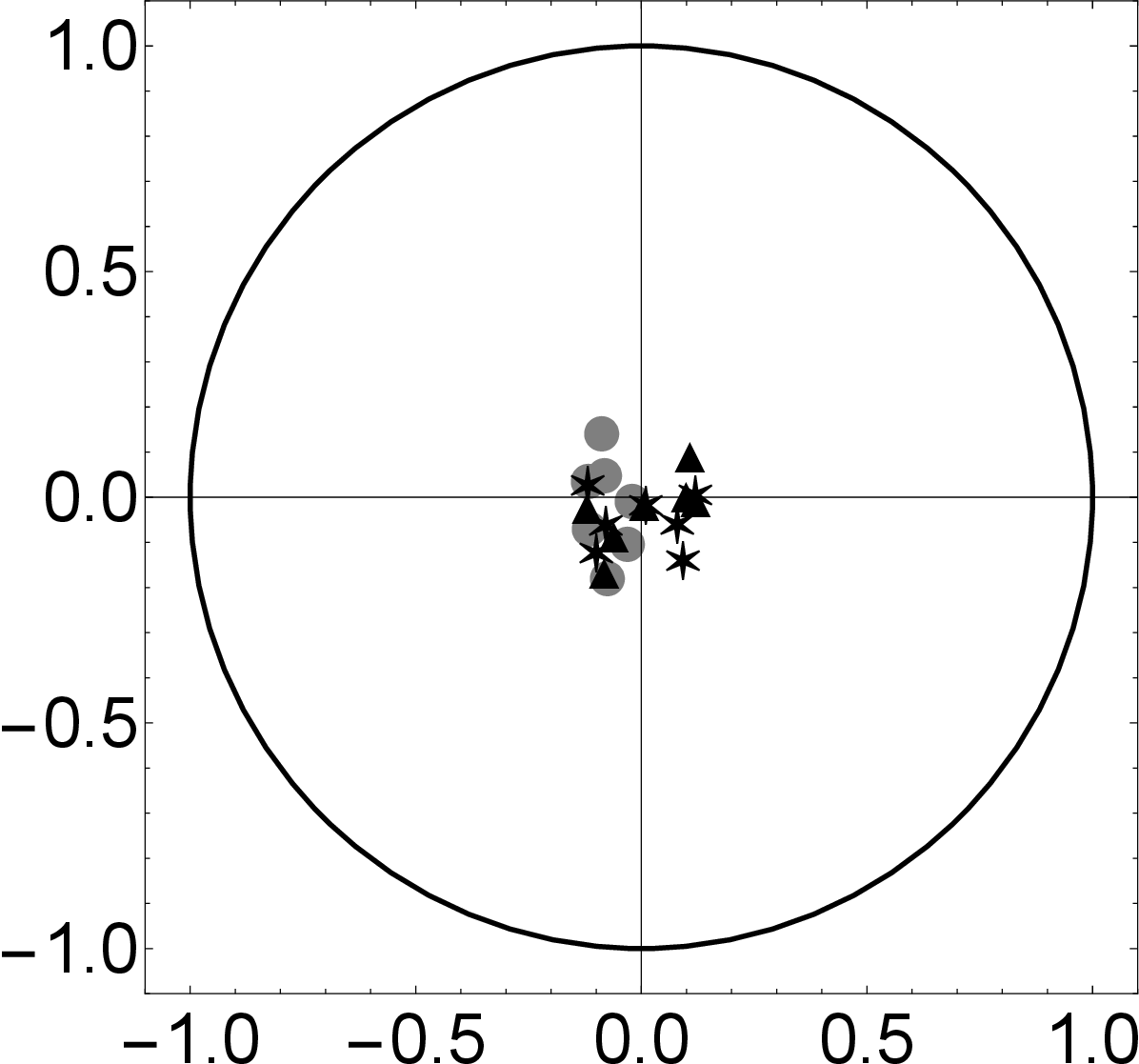}&\includegraphics[width=0.32\textwidth]{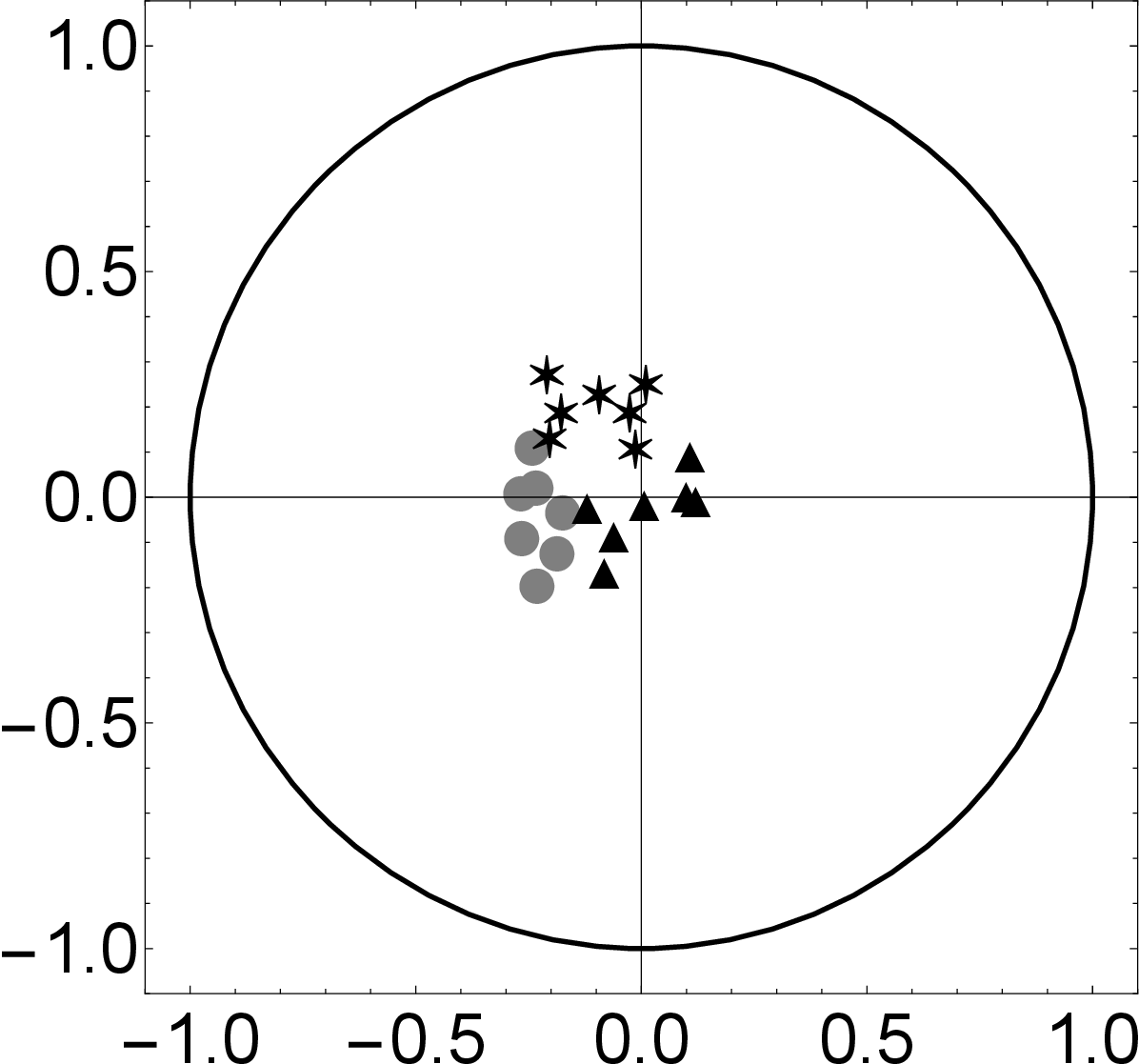}&\includegraphics[width=0.32\textwidth]{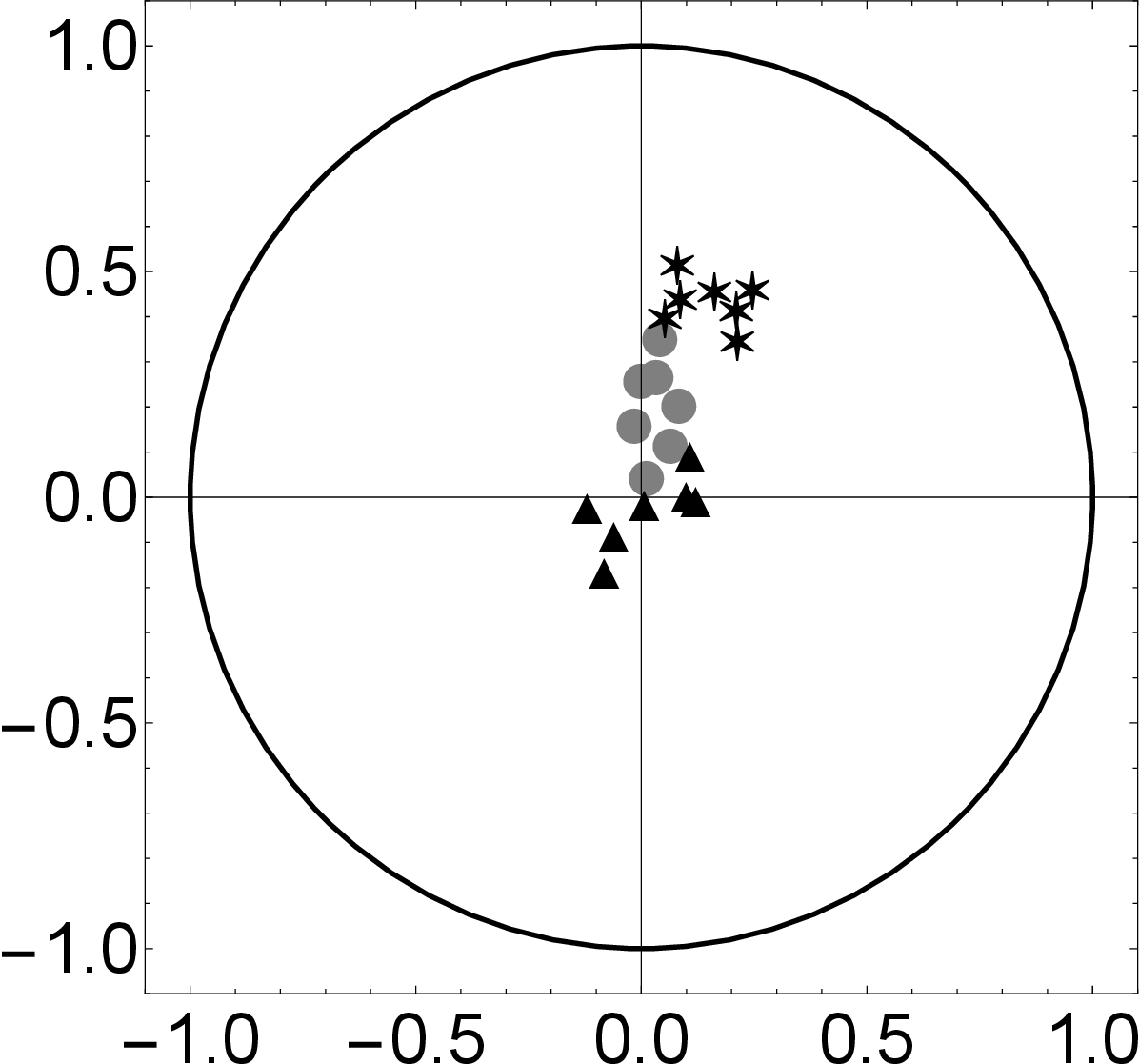}\\
    a)&b)&c)
  \end{tabular}
  \caption{\label{fig7}Position of points at the moments T = 0, 2, 4.}
\end{figure*}

Initial embeddings for three trees are shown in Figure \ref{fig7}a). These embeddings are proper for each tree viewed separately (i.e. if we ignore intra-layer links). The threshold for hyperbolic embedding is set at $\epsilon=1/5$. Denote by $A_i, B_j$ and $C_k$ the nodes of trees $A,B$ and $C$, respectively. Further, denote by $z_i$ points in the unit disc, corresponding to the initial embeddings of nodes $A_i$, by $\xi_j$ points corresponding to initial embeddings of nodes $B_j$ and by $\zeta_k$ the same for $C_k$.

Denote by $E_{A,B}$ the set of intra-layer connections between the nodes $A_i$ and $B_j$ and define $E_{A,C}$ and $E_{B,C}$ in analogous way. From Figure \ref{fig2} it is apparent that $(A_6,B_7) \in E_{A,B}$; $(A_2,C_6), (A_5,C_1), (A_5,C_3) \in E_{A,C}$, while $E_{B,C}$ is the empty set.

Further, apply conformal transformations (isometries of the hyperbolic disc) $g^A$ and $g^B$ to points $z_i$ and $\xi_j$, respectively.

Define the total error function as follows:
$$
J(g^A,g^B) = \sum \limits_{i,j} h^{(A,B)}(g^A(z_i),g^B(\xi_j)) + \sum \limits_{i,k} h^{(A,C)}(g^A(z_i),\zeta_k)) + \sum \limits_{j,k} h^{(B,C)}(g^B(\xi_j),\zeta_k),
$$
where the functions $h$ are given by:
\[
h^{(A,B)}(\alpha,\beta) = \begin{cases}
  1 & \mbox{ if $ d_{hyp}(\alpha,\beta) < \epsilon $ and $(A_i,B_j) \notin E_{A,B}$}\\
  1 & \mbox { if $ d_{hyp}(\alpha,\beta) > \epsilon $ and $(A_i,B_j) \in E_{A,B}$} \\
    0 & \mbox{ if $ d_{hyp}(\alpha,\beta) > \epsilon $ and $(A_i,B_j) \notin E_{A,B} $}\\
  -1 & \mbox{ if $ d_{hyp}(\alpha,\beta) < \epsilon $ and $(A_i,B_j) \in E_{A,B} $}
  \end{cases}
  \]
  and analogously for $h^{(A,C)}$ and $h^{(B,C)}$.

The above definition means that errors of both types (mapping two unconnected nodes from different layers to points at the distance less than $\epsilon$, or mapping two connected nodes from different layers to points at the distance greater than $\epsilon$) are penalized by $1$. On the other hand, exact scores of the kind that two connected points are mapped into points at the distance below $\epsilon$ are awarded by $-1$.

In such a setup, the minimal value of the function, corresponding to the perfect embedding is equal to the number of intra-layer connections (in our case it is $-4$, corresponding to the four dashed lines in Figure \ref{fig2}).

Our goal is to find two M\" obius transformations of the unit disc that minimize the error function $J(g^A,g^B)$. The problem is formalized as:
$$
\mbox{ minimize }  J(g^A,g^B), \mbox{ where } g^A,g^B \in G_+.
$$
This is an optimization problem over the Lie group $G_+ \times G_+$.

In order to learn coupled isometries of $\mathbb{B}^2$ we simulated the dynamics \eqref{sub_swarms} for $p=2$ sub-swarms at the time interval for $t \in [0,4]$. Notice that we can leave embedding of one tree intact, and try to adjust embeddings of the other two trees.

We trained the model \eqref{sub_swarms} with the objective of minimizing the error function.

\begin{figure*}[h]
\centering
  \begin{tabular}{@{}c@{}}
    \includegraphics[width=0.5\textwidth]{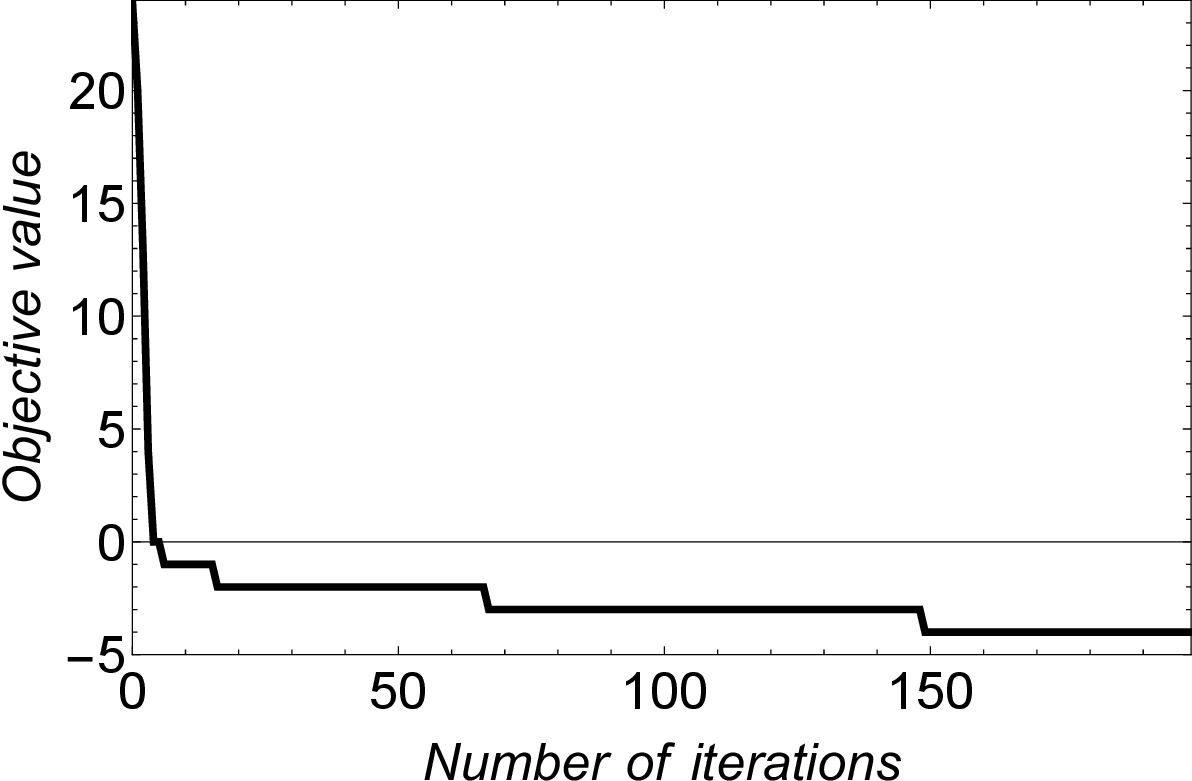}\\
  \end{tabular}
  \caption{\label{fig8}Evolution of the error function.}
\end{figure*}

In Figure \ref{fig7} we depict initial embeddings, intermediate embeddings at a certain the middle of algorithm and the final configuration in Figure \ref{fig7}c). Figure \ref{fig8} shows the evolution of the error function.

Underline that this problem does not always has a solution. Existence of the solution depends on initial embeddings and specific intra-layer links.  In some simulations our algorithm was not able to find two isometries that minimize the error to zero, most likely because such transformations did not exist.

\subsection{Exploring directional labyrinth}

The problem of exploring directional labyrinth boils down to learning $10$ angles, i.e. ten points on the unit circle $\mathbb{S}^1$.
One model that can encode this kind of data is the famous Kuramoto model of coupled phase oscillators \cite{Kuramoto}.

\noindent{\bf Method 1.}

We use the Kuramoto model with ten oscillators of the following form:
$$
\dot \varphi_j = \omega + K_{j-1,j} \sin (\varphi_{j-1} - \varphi_j) + K_{j,j+1} \sin (\varphi_{j+1}-\varphi_j), j=2,\dots,9
$$
and
$$
\dot \varphi_1 = \omega + K_{12} \sin (\varphi_2 - \varphi_1), \quad \dot \varphi_{10} = \omega + K_{9,10} \sin (\varphi_9 - \varphi_{10}).
$$
Hence, we assume that oscillators are coupled through an array. Training of such a model consists in learning ten parameters $\omega,K_{12},K_{23},\dots,K_{9,10}$ with the objective of maximizing total reward.

We simulated the above Kuramoto dynamics with the initial points initially sampled from the uniform distribution on the unit circle. The parameters $\omega, K_{j-1,j}$ are updated using the CMA-ES algorithm.

\begin{figure*}[h]
\centering
  \begin{tabular}{@{}cc@{}}
    \includegraphics[width=0.5\textwidth]{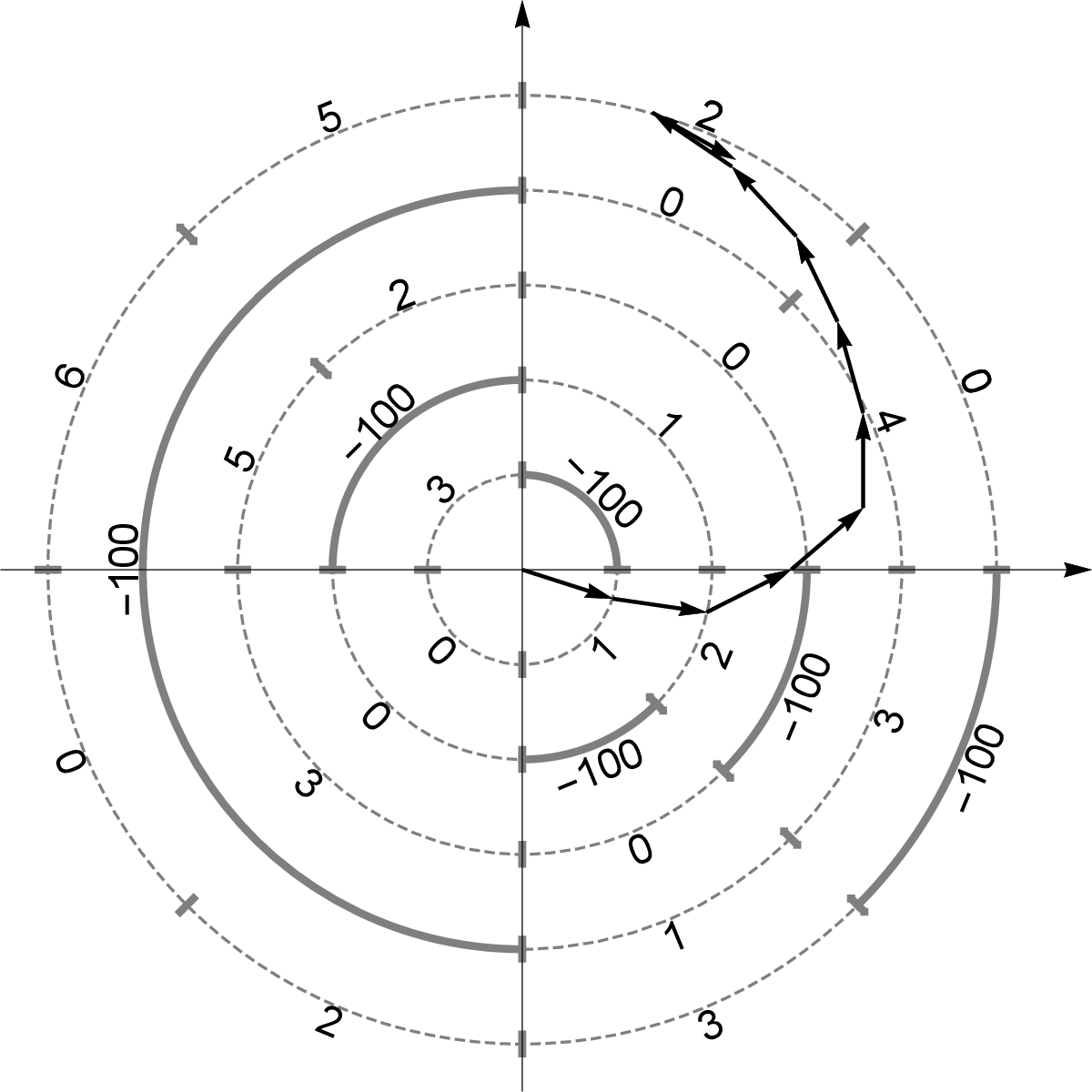}&\includegraphics[width=0.5\textwidth]{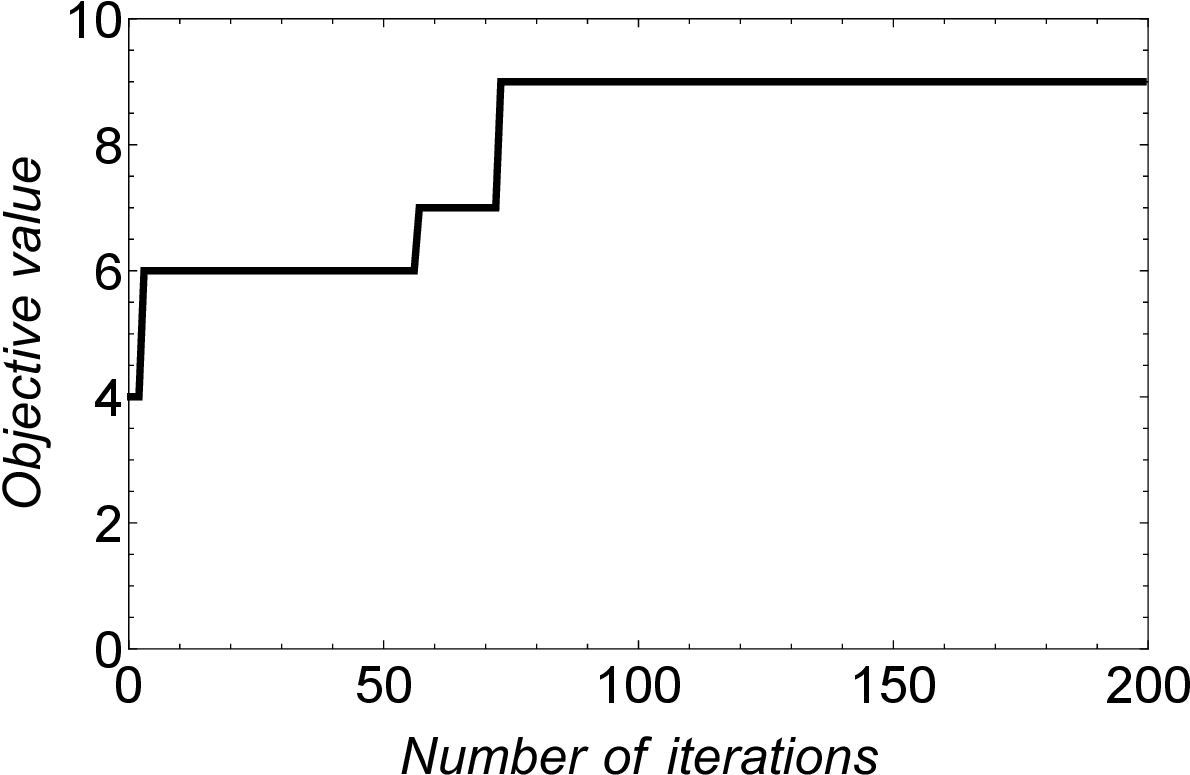}\\
    a)&b)
  \end{tabular}
  \caption{\label{fig9} Simulation results of the problem 4.3 with method 1: a) agent's path, and b) evolution of the reward function.}
\end{figure*}

The algorithm learned the path shown in Figure \ref{fig9}a) and evolution of the total reward in Figure \ref{fig9}b).

\noindent{\bf Method 2}

Another option is to implement stochastic policies for learning $10$ angles $\varphi_1,\dots,\varphi_{10}$. We encode the stochastic policies with ten wrapped Cauchy distributions \eqref{wrapped_Cauchy}. In their turn, these distributions are parametrized by points $a_1,\dots,a_{10}$ in the unit disc. Hence, the stochastic policy is determined by a configuration of $10$ points in the hyperbolic unit disc.
\begin{enumerate}
\item[i)] Randomly sample $10$ points $a_{1}(0), ..., a_{10}(0)$ in $\mathbb{B}^2$.

\item[ii)] Introduce the Poincar\' e swarm \eqref{swarm} with $N=10$ complex oscillators and nine couplings $K_{12},\dots,K_{9,10}$. An additional parameter is $\omega \geq 0$. Phase-shifts $\beta_{jk}$ are not included in the model, as well as interactions $K_{jk}$ for $k-j \neq \pm 1$.

\item[iii)] Run the swarming dynamics defined in ii) over the time interval $[0, 1]$, starting from the initial conditions $a_{1}(0), ..., a_{10}(0).$

\item[iv)] For each evolved point $a_{i}(1)$, sample $5$ angles from the wrapped Cauchy distribution \eqref{wrapped_Cauchy} with the parameter $a = a_i(1)$. (Sampling from such a distribution is easy: we first sample a uniformly distributed point on $\mathbb{S}^1$ and transform it by the mapping \eqref{Mobius} with $a=a_i(1)$.)

\item[v)] We have $5$ sequences of $10$ sampled angles (points on the circle). For each of them construct the path through the labyrinth and calculate the average total reward.

\item[vi)] Train the Poincar\' e swarm, i.e. update couplings $K_{j-1,j}$ and $\omega$ using CMA-ES.

\item[vii)] Repeat the steps ii)-iv) until the total reward stabilize over simulations.
\end{enumerate}

\begin{figure*}[h]
\centering
  \begin{tabular}{@{}cc@{}}
    \includegraphics[width=0.5\textwidth]{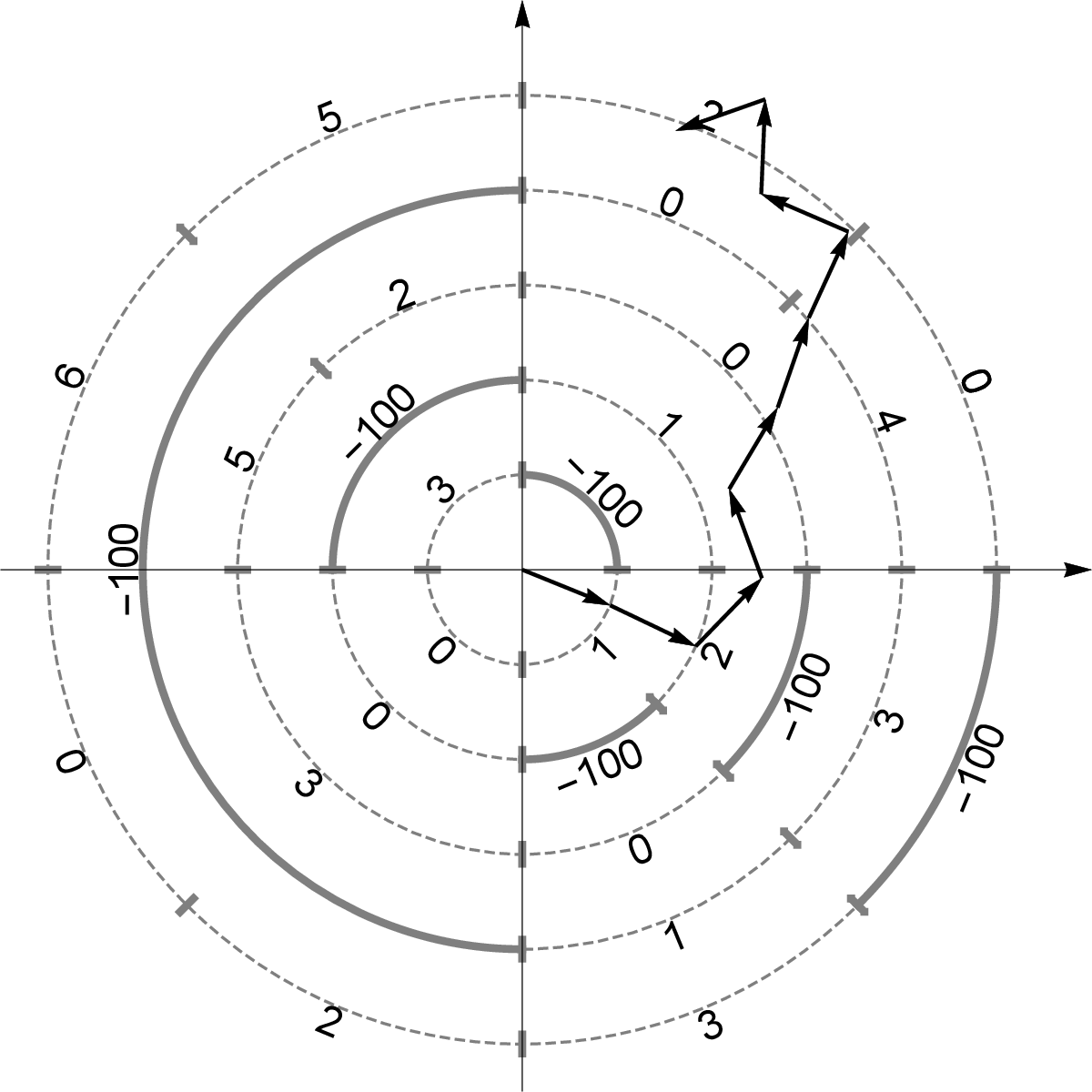}&\includegraphics[width=0.5\textwidth]{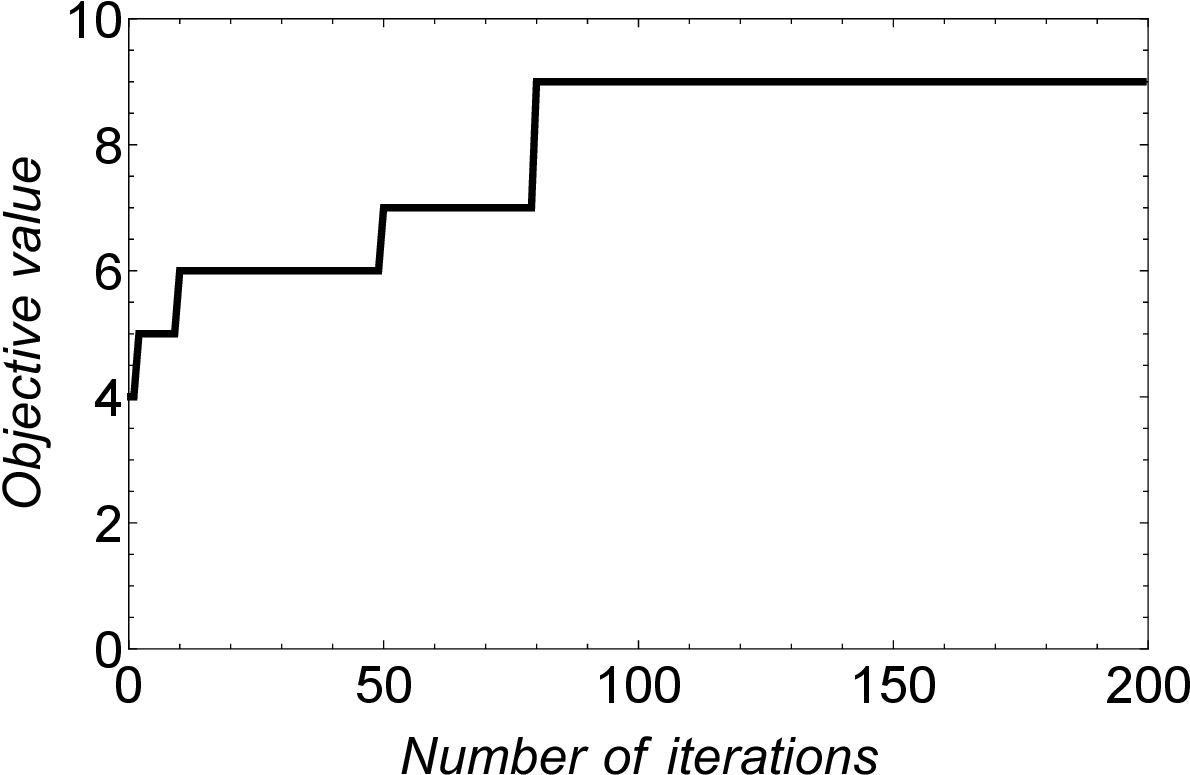}\\
    a)&b)
  \end{tabular}
  \caption{\label{fig10} Simulation results of the problem 4.3 with method 2: a) agent's path, and b) evolution of the reward function.}
\end{figure*}

\begin{figure*}[h]
\centering
  \begin{tabular}{@{}ccc@{}}
    \includegraphics[width=0.32\textwidth]{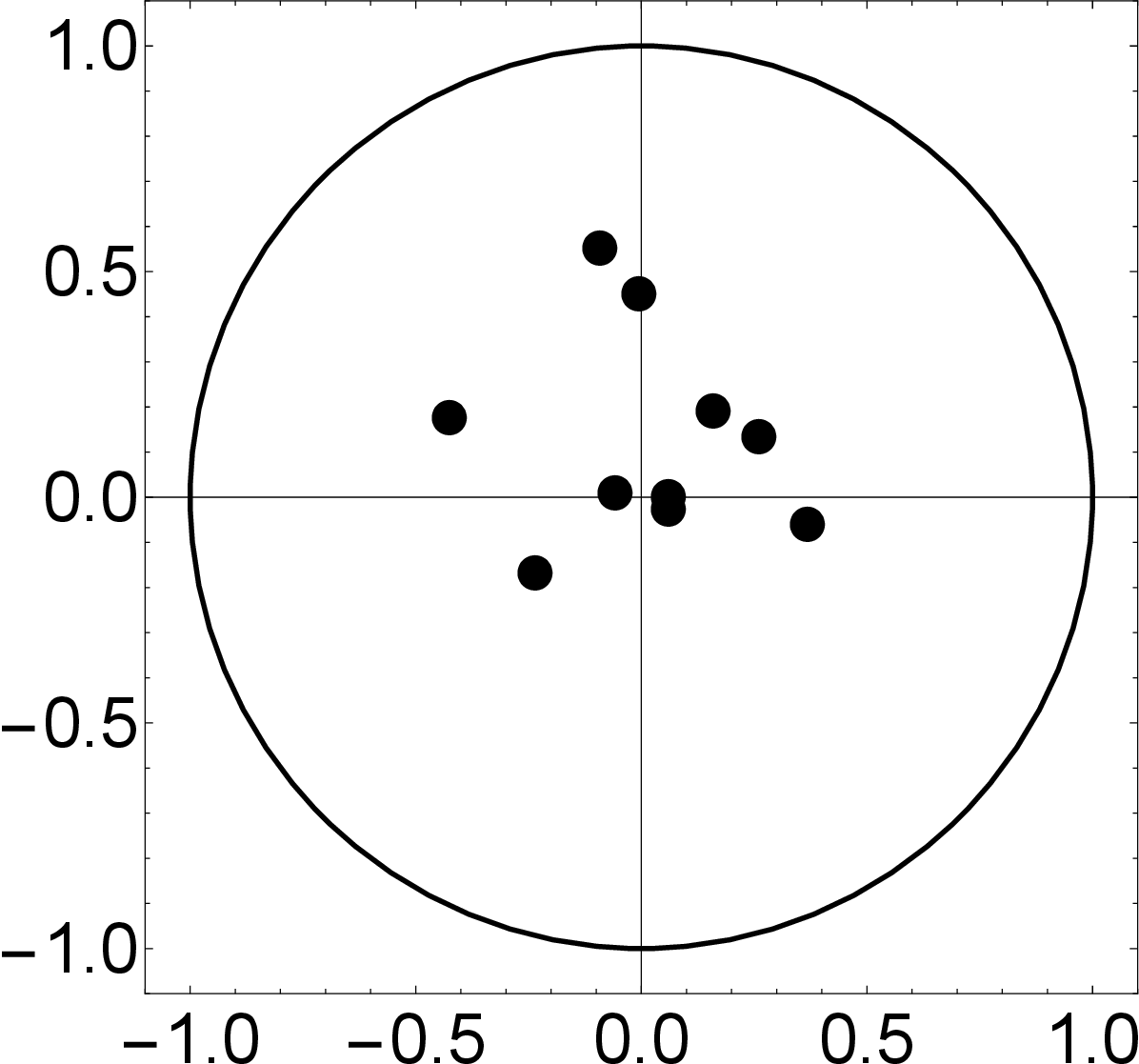}&\includegraphics[width=0.32\textwidth]{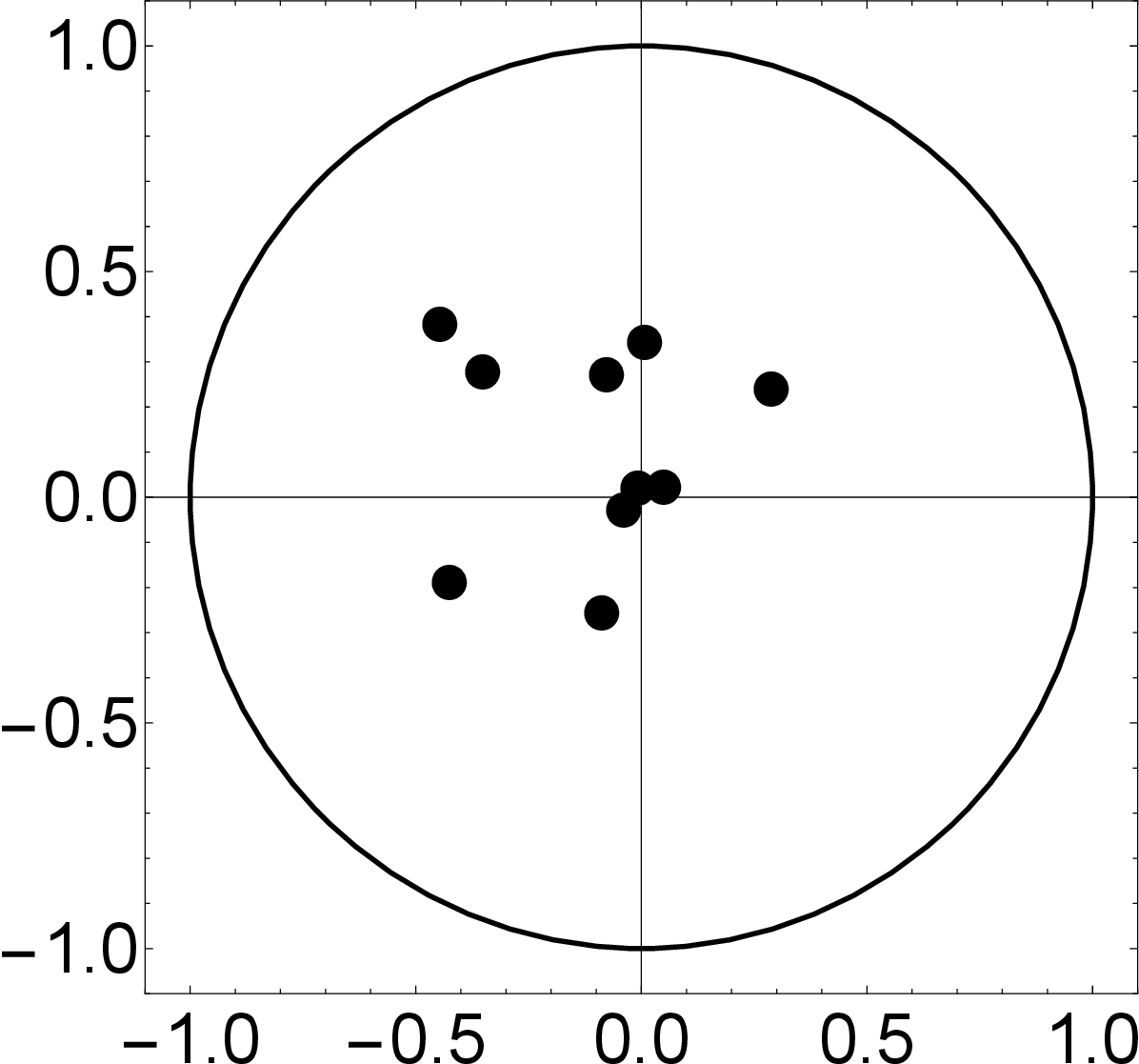}&\includegraphics[width=0.32\textwidth]{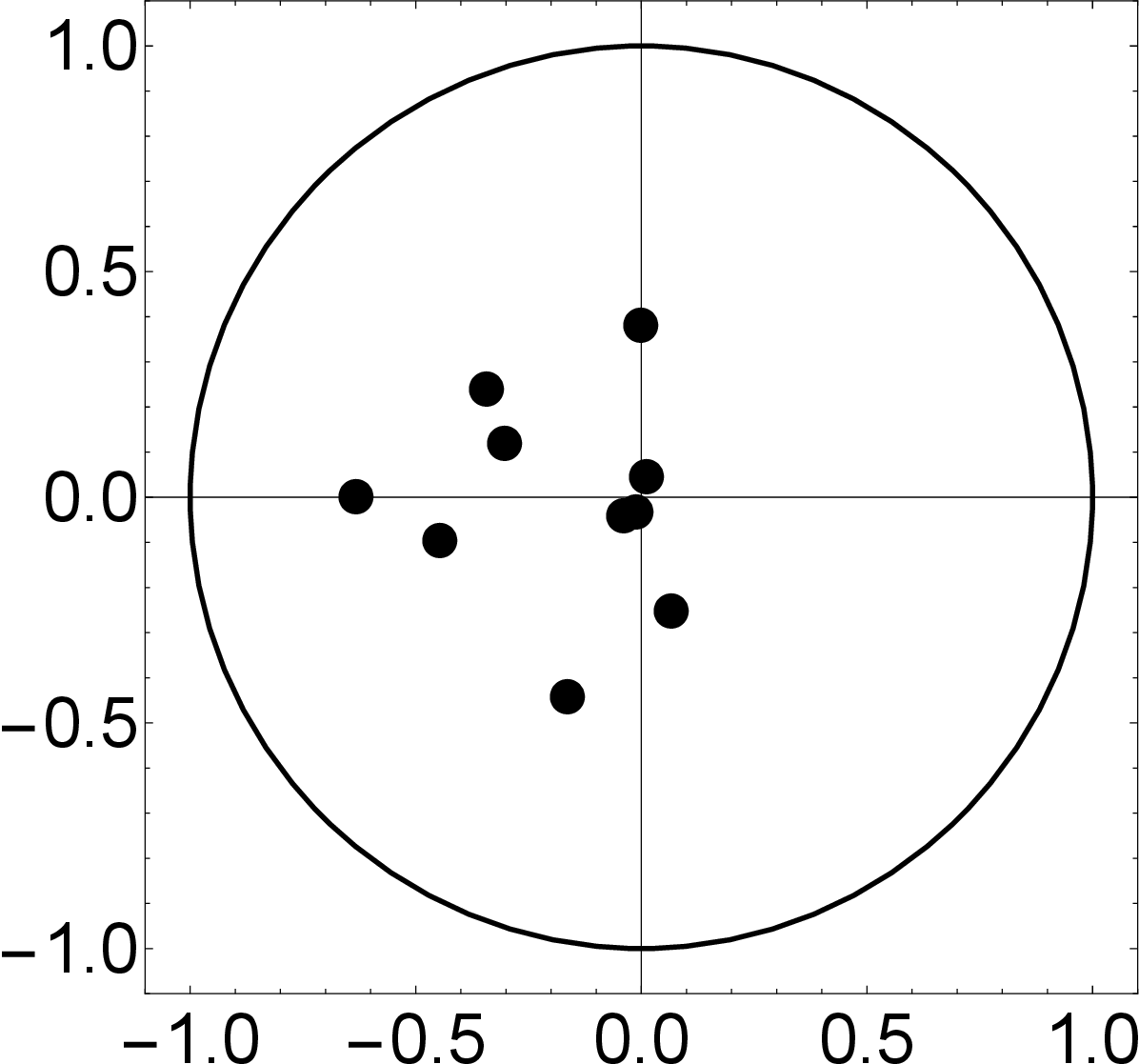}\\
    a)&b)&c)
  \end{tabular}
  \caption{\label{fig11}Position of points at the moments $T=0,0.5,1$.}
\end{figure*}

A typical path path sampled from the stochastic policy is shown in Figure \ref{fig10}a) and the evolution of total reward in Figure \ref{fig10}b). Evolution of parameters $a_i$ is shown in Figure \ref{fig11}. Notice that algorithm preserves a significant degree of randomness, as points $a_i$ do not fully approach the circle, see Figure \ref{fig11}c).


\subsection{Two frogs}

In this simulation we assume from the beginning that players will make 3 jumps each. The policies are learned by two swarms \eqref{swarm} with 3 complex oscillators each, which do not interact directly. Instead, they are indirectly coupled through the reward. We implement the following algorithm:
\begin{enumerate}
\item[i)] Sample 6 points in the hyperbolic disk $\mathbb{B}^2$ from \eqref{JacMark} with $a=0$ and $s=2$.

\item[ii)] Transform each of these points to parameters of normal distributions $(m_1, \sigma_{1}^{2}),...,(m_6, \sigma_{6}^{2})$.

\item[iii)] Sample random numbers from the distributions obtained in step ii) for both players. These numbers represent the lengths of jumps.

\item[iv)] Repeat steps iii) and iv) 10 times, and calculate average rewards for both players.

\item[v)] Use the 6 initial points as initial conditions for two Poincar\' e swarms with 3 complex oscillators each.

\item[vi)] Train the Poincar\' e swarms.

\item[vii)] Repeat the steps ii)-iv) until the reward of both players stabilize over the time.
\end{enumerate}

\begin{figure*}[h]
\centering
  \begin{tabular}{@{}c@{}}
    \includegraphics[width=0.5\textwidth]{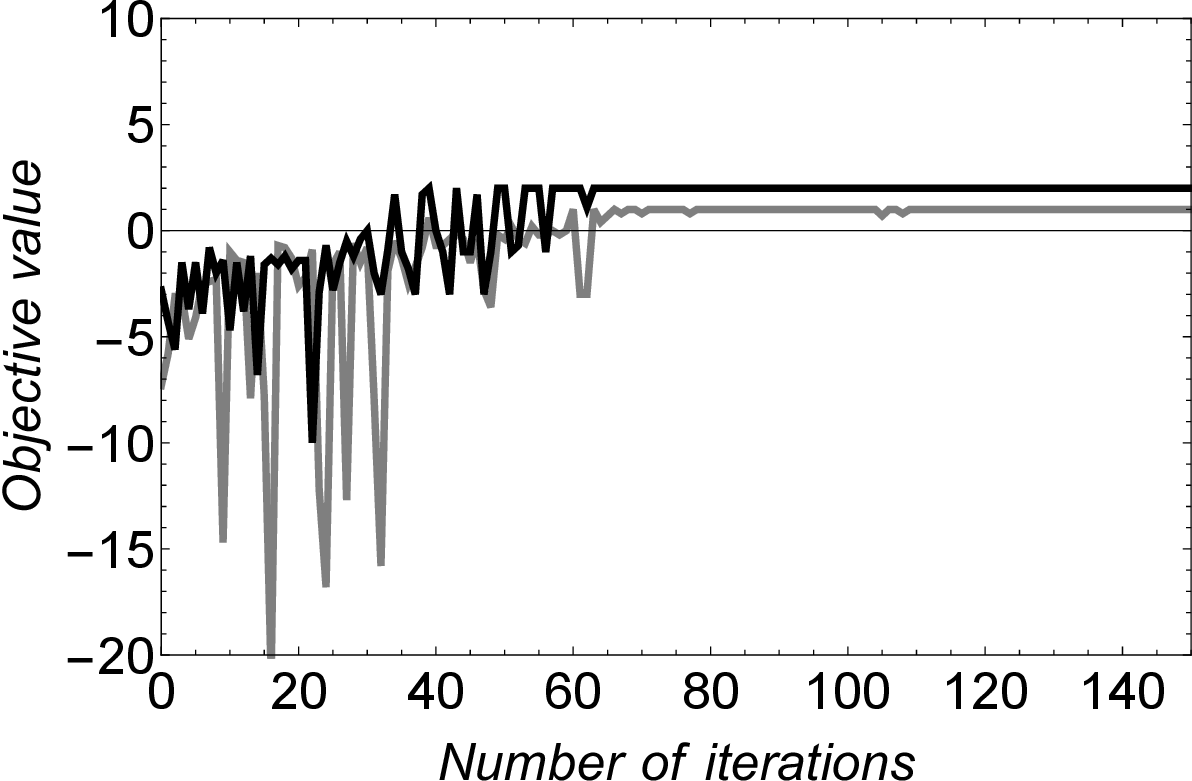}\\
  \end{tabular}
  \caption{\label{fig12} Evolution of the reward function: player A (black line) and player B (gray line).}
\end{figure*}

\begin{figure*}[h]
\centering
  \begin{tabular}{@{}ccc@{}}
    \includegraphics[width=0.32\textwidth]{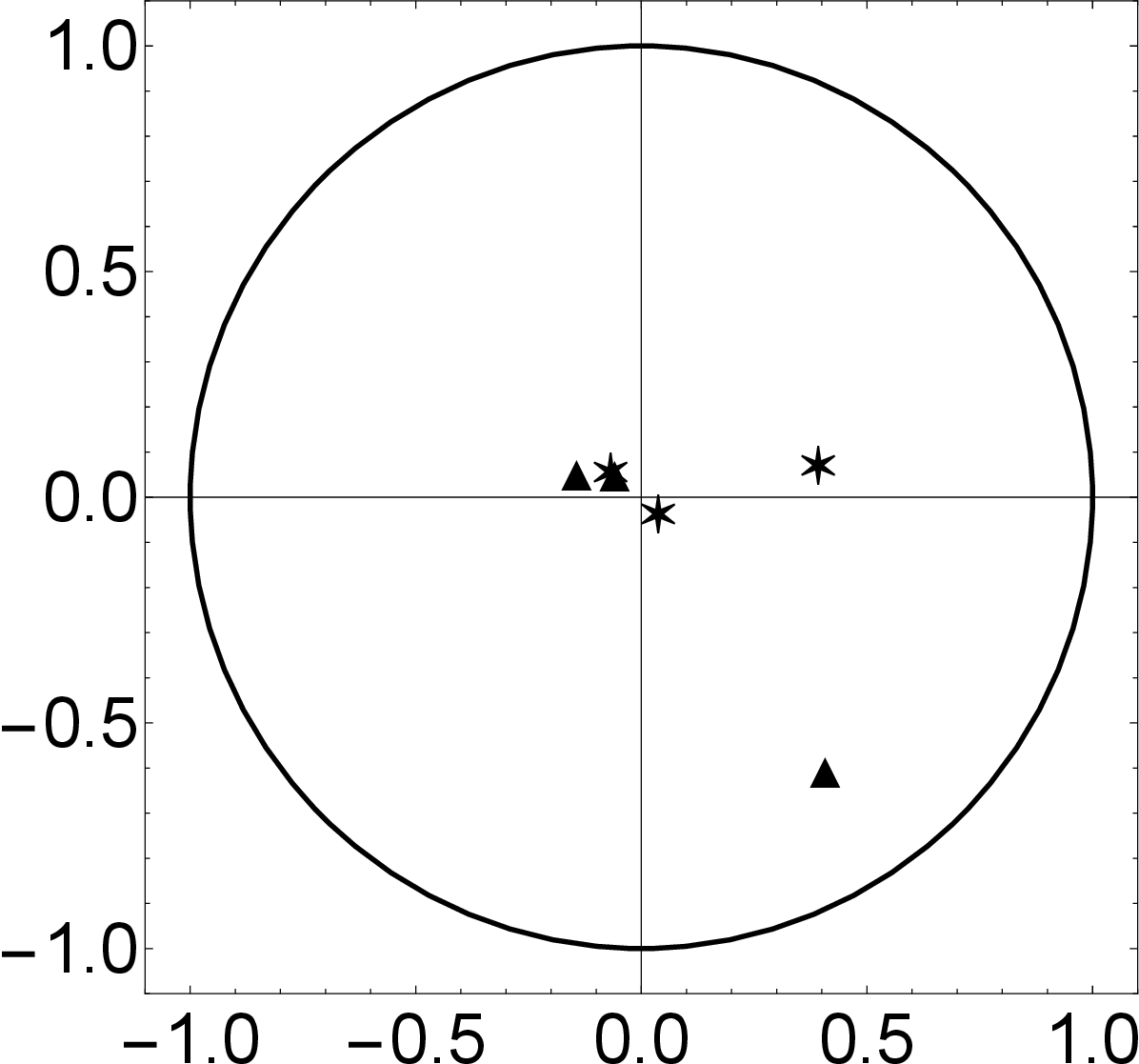}&\includegraphics[width=0.32\textwidth]{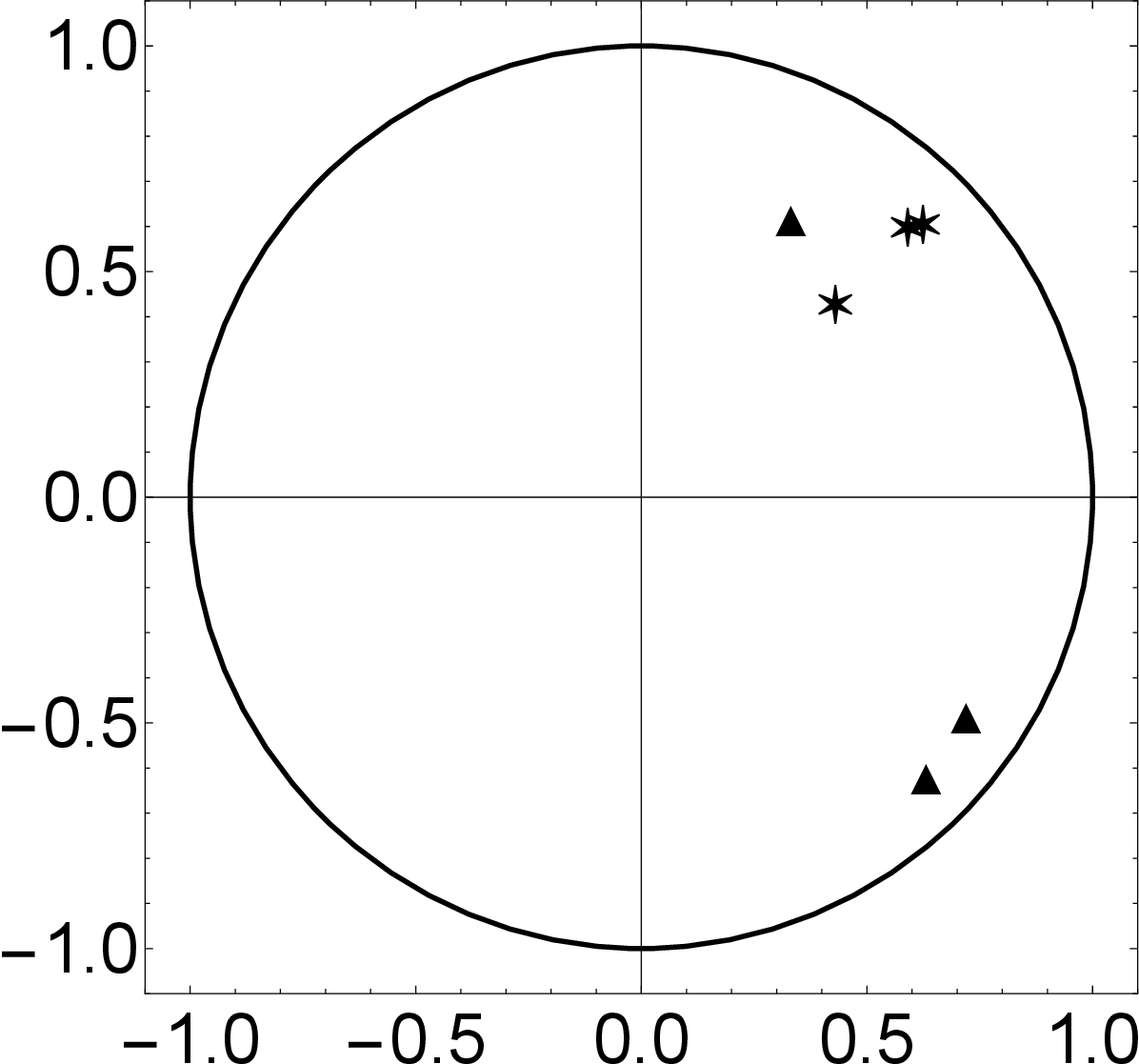}&\includegraphics[width=0.32\textwidth]{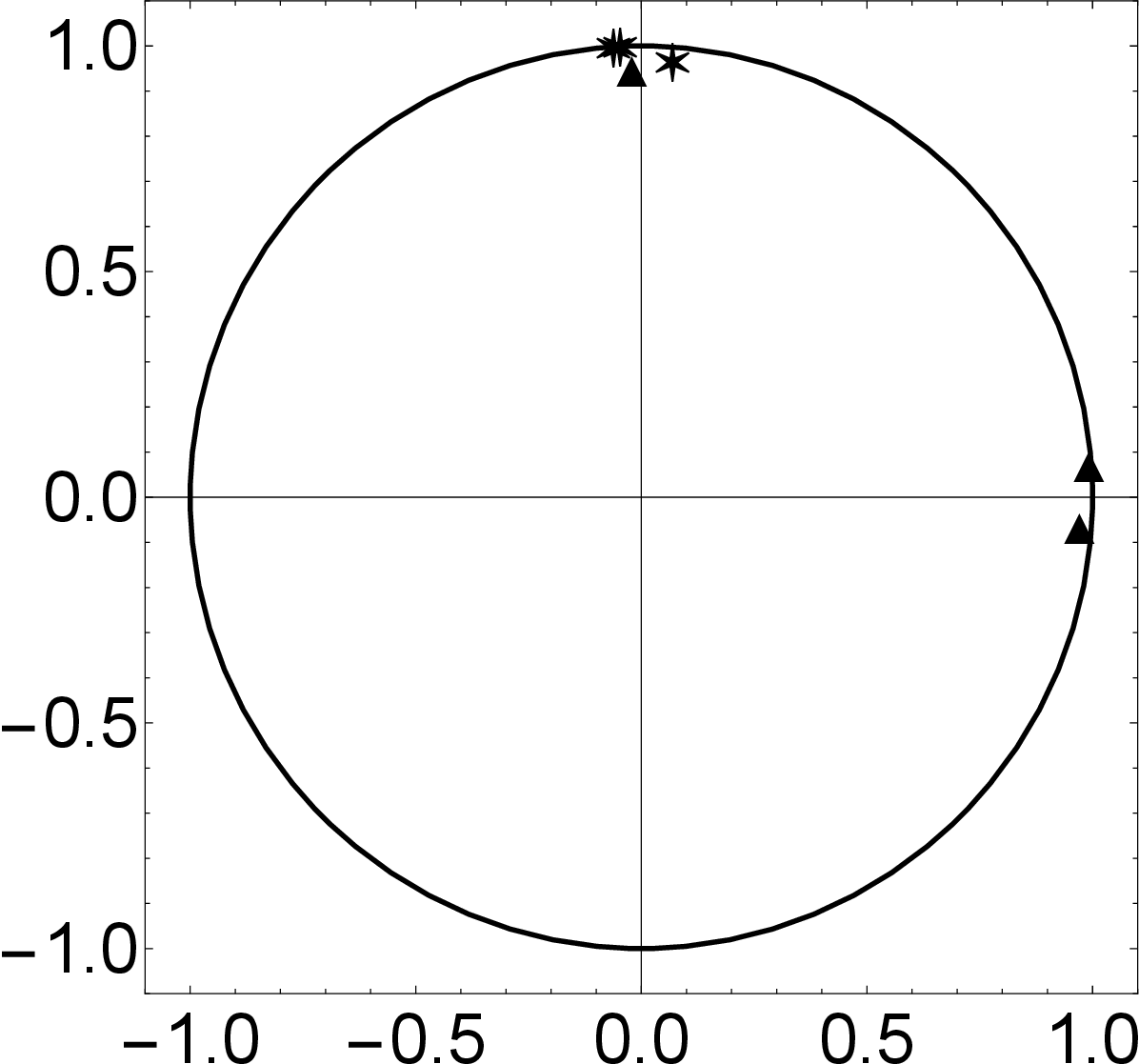}\\
    a)&b)&c)
  \end{tabular}
  \caption{\label{fig13}Position of points at the moments $T=0,1,2$ with corresponding distributions: a) ${\cal N}(0.07, 0.31)$, ${\cal N}(-0.07, 0.65)$, ${\cal N}(0.13, 0.81)$, ${\cal N}(0.16, 0.94)$, ${\cal N}(0.13, 0.79)$, ${\cal N}(-0.51, 0.14)$; b) ${\cal N}(0.40, 0.058)$, ${\cal N}(0.41, 0.07)$, ${\cal N}(0.38, 0.20)$, ${\cal N}(-0.30, 0.06)$, ${\cal N}(0.58, 0.16)$, ${\cal N}(-0.41, 0.05)$, and c) ${\cal N}(1.05, 0.001)$, ${\cal N}(1.06, 0.001)$, ${\cal N}(0.93, 0.02)$, ${\cal N}(0.04, 0.002)$, ${\cal N}(1.02, 0.03)$, ${\cal N}(-0.03 0.01)$. }
\end{figure*}

Evolutions of the rewards for player A and player B are shown in Figure \ref{fig12}. The algorithm ends with one player agreeing on a lower reward than optimal. The typical scenario is when the first player makes the jumps of length 1,2 and 2, while the second 3,2 and -4. Evolution of strategies towards deterministic ones is presented in Figure \ref{fig13}.


\subsection{Two agents exploring environment in the plane}\label{two_agent_pl}

We assume that both players implement stochastic policies encoded by three one-variate normal distributions. In other words, each of them learns the set of parameters $(m_1,\sigma_1),(m_2,\sigma_2),(m_3,\sigma_3)$. We use the same model with two uncoupled Poincar\' e swarms as in the previous example.
\begin{enumerate}
\item[i)] Sample 6 points in the hyperbolic disk $\mathbb{B}^2$ from \eqref{JacMark} with $a=0$ and $s=2$.

\item[ii)] Transform each of these points to parameters of normal distributions $(m_1, \sigma_{1}^{2}),...,(m_6, \sigma_{6}^{2})$.

\item[iii)] Sample random numbers $a_1,a_2,a_3$ and $b_1,b_2,b_3$ from the distributions obtained in step ii) for both players and construct vectors $(a_1,b_1),(a_2,b_2),(a_3,b_3)$.

\item[iv)] Repeat steps iii) and iv) 10 times, and calculate average reward.

\item[v)] Use the 6 initial points as initial conditions for two Poincar\' e swarms \eqref{swarm} each of them consisting of 3 complex oscillators.

\item[vi)] Train the Poincar\' e swarms using the CMA-ES algorithm.

\item[vii)] Repeat the steps ii)-iv) until the reward of both players stabilize over the time.
\end{enumerate}
\begin{figure*}[h]
\centering
  \begin{tabular}{@{}cc@{}}
   \includegraphics[width=0.5\textwidth]{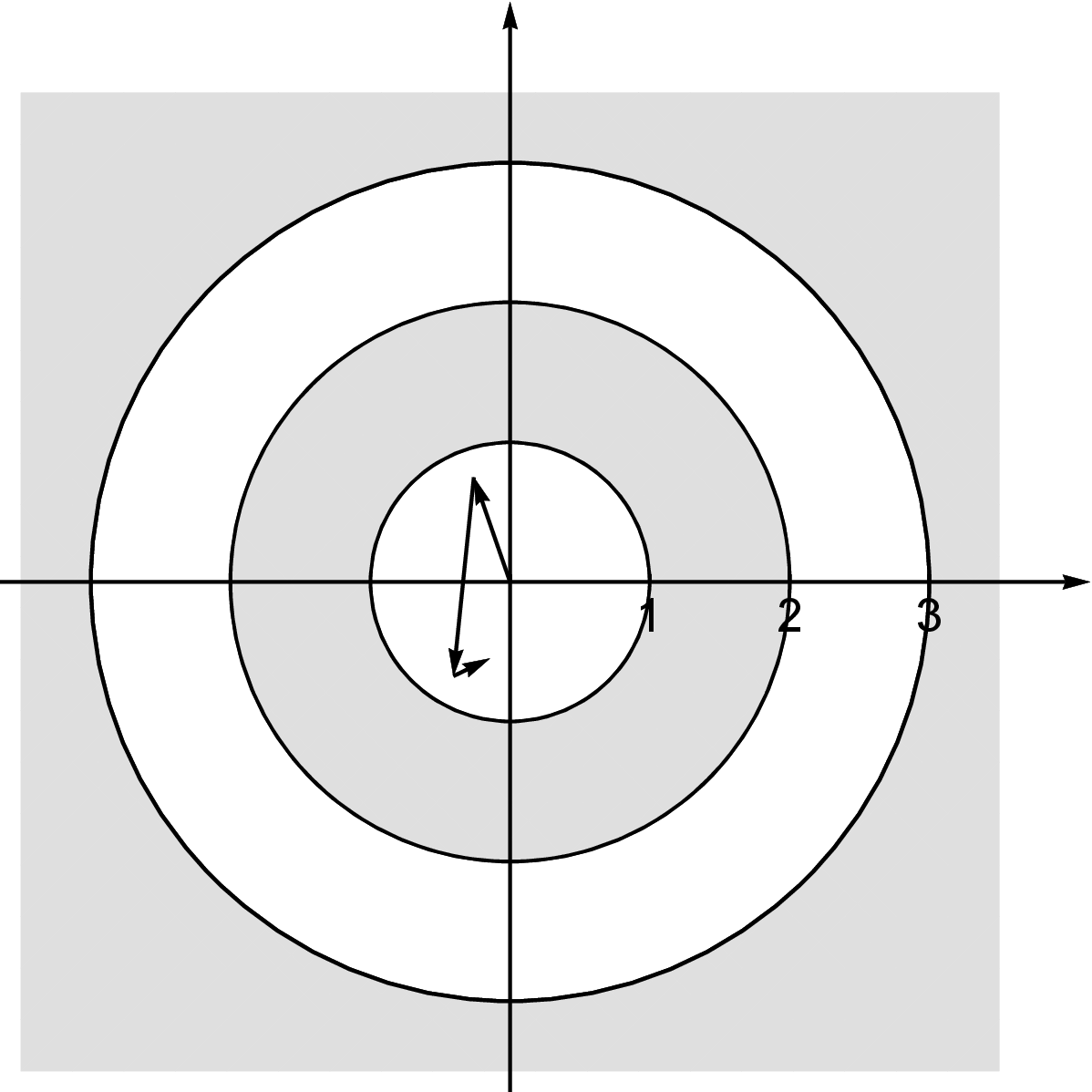}&\includegraphics[width=0.5\textwidth]{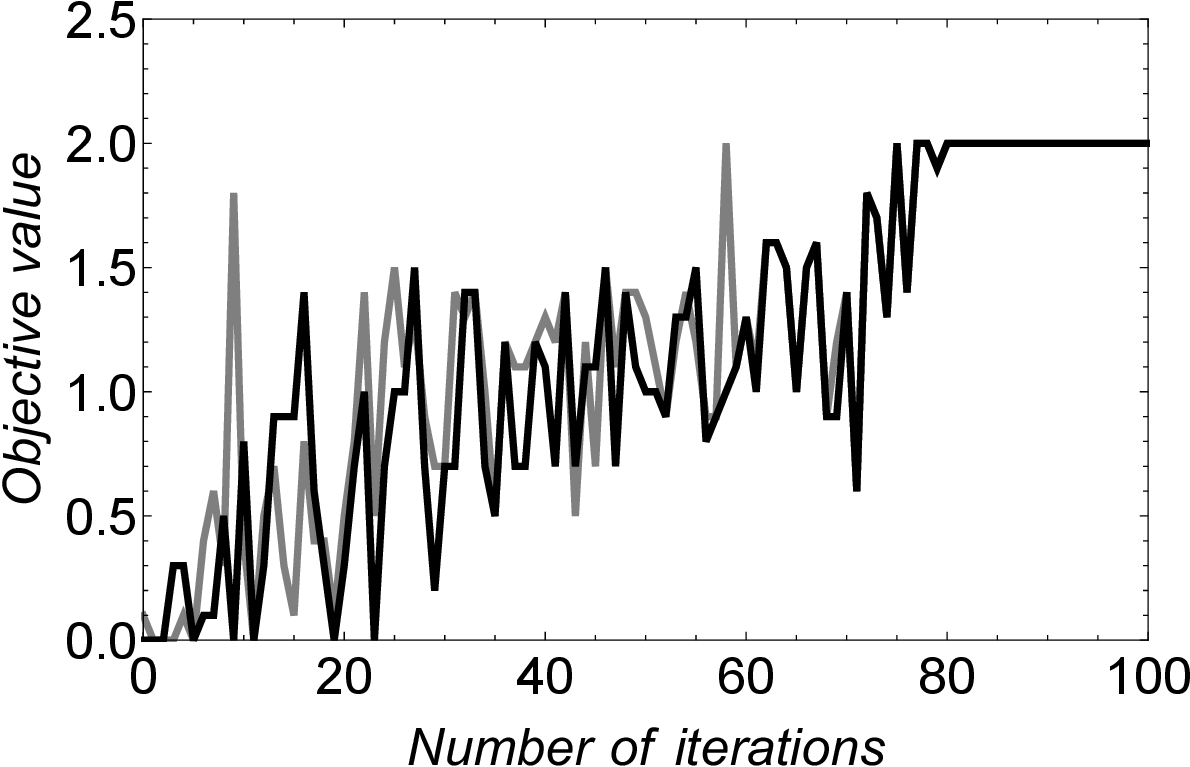}\\
   a) & b)
  \end{tabular}
  \caption{\label{fig14} Simulation results of the problem \ref{two_agent_pl}: a) path, and b) evolution of average rewards for player A (black line) and player B (gray line).}
\end{figure*}
\begin{figure*}[h]
\centering
  \begin{tabular}{@{}ccc@{}}
    \includegraphics[width=0.32\textwidth]{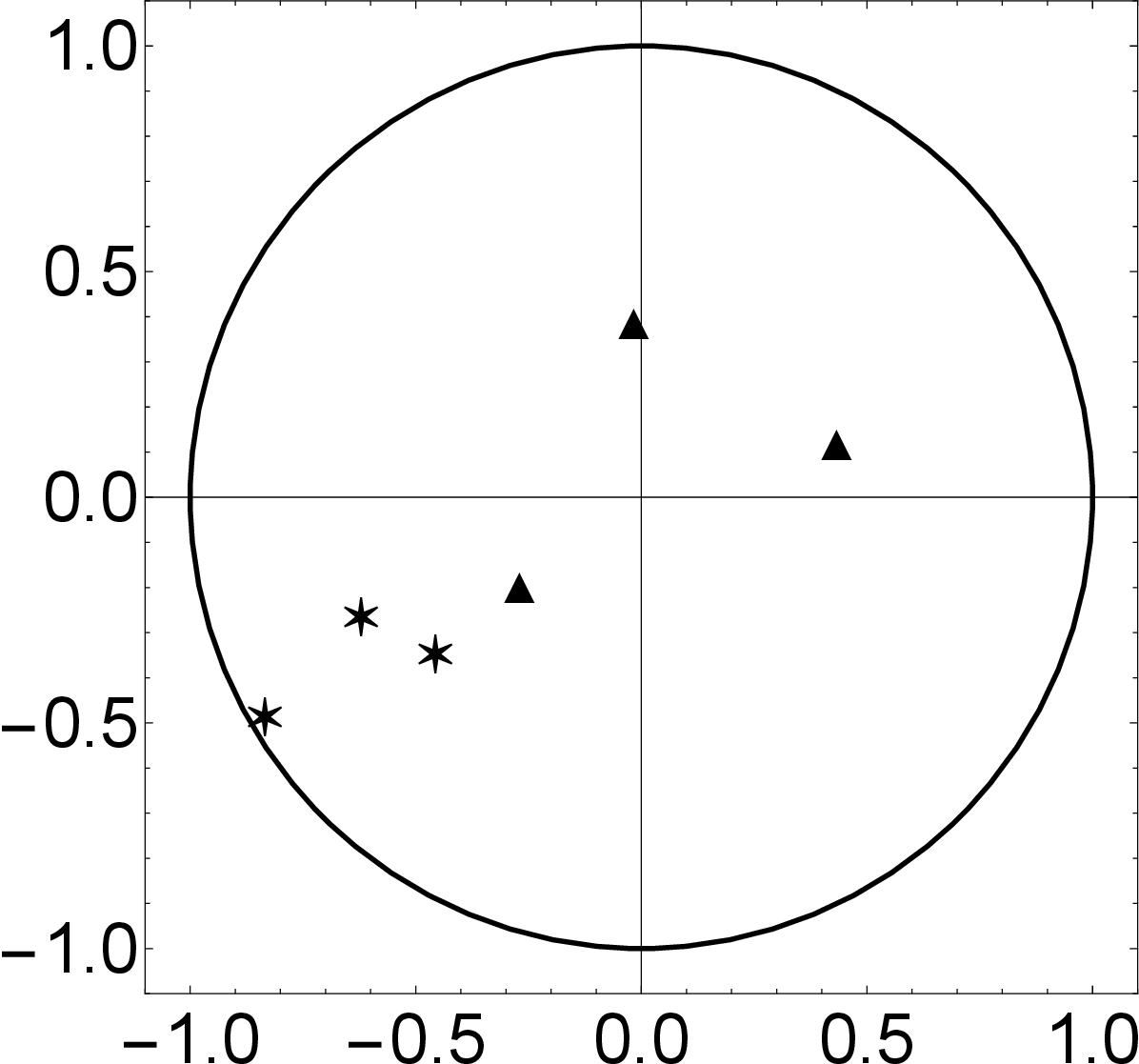}&\includegraphics[width=0.32\textwidth]{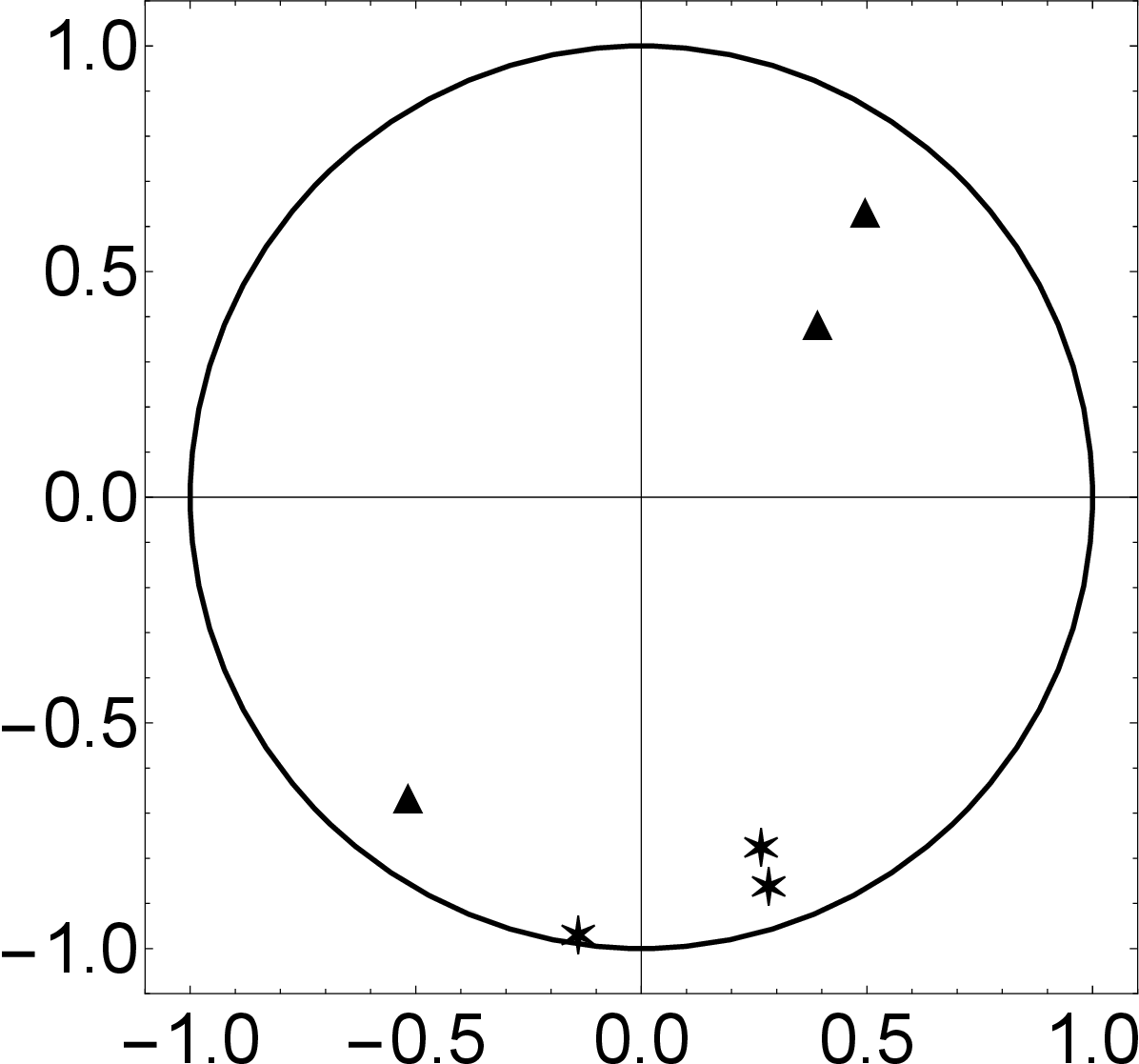}&\includegraphics[width=0.32\textwidth]{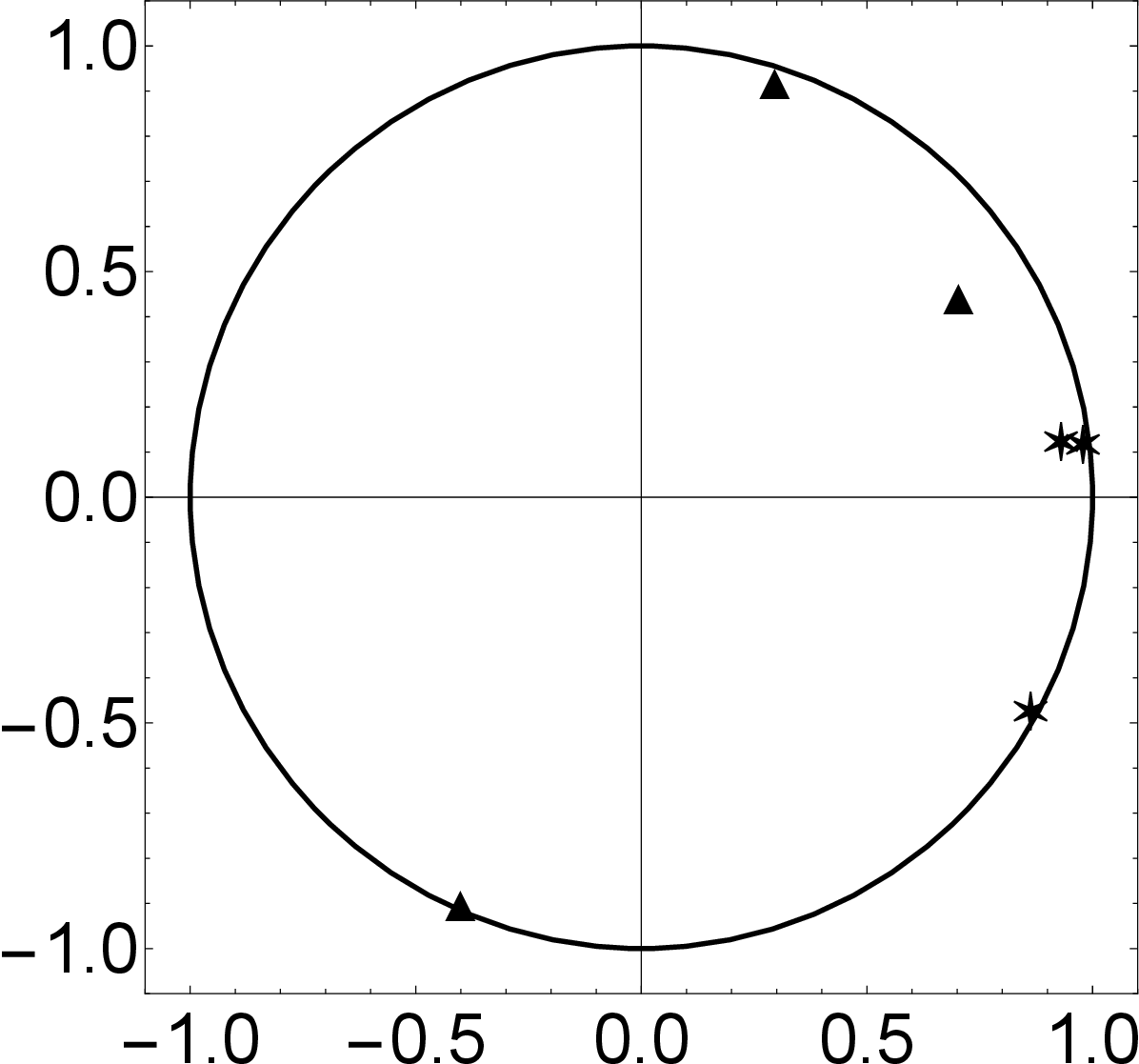}\\
    a)&b)&c)
  \end{tabular}
  \caption{\label{fig15}Position of points at the moments $T=0,1,2$ with corresponding distributions: a) ${\cal N}(-3.69, 0.18)$, ${\cal N}(-1.67, 1.15)$, ${\cal N}(-2.48, 1.81)$, ${\cal N}(0.12, 0.27)$, ${\cal N}(-0.67, 1.11)$, ${\cal N}(0.70, 0.53)$; b) ${\cal N}(-1.15, 0.02)$, ${\cal N}(-0.70, 0.11)$, ${\cal N}(-0.72, 0.05)$, ${\cal N}(0.48, 0.09)$, ${\cal N}(-1.97, 0.32)$, ${\cal N}(0.38, 0.24)$, and c) ${\cal N}(-0.26, 0.01)$, ${\cal N}(0.07, 0.02)$, ${\cal N}(0.06, 0.01)$, ${\cal N}(0.73, 0.02)$, ${\cal N}(-1.54, 0.02)$, ${\cal N}(0.29, 0.07)$. }
\end{figure*}

Evolution of rewards and strategies are shown in Figure \ref{fig14}b) and \ref{fig15}, repsectively. We see that the algorithm ends up with finding a compromise at the field $(2,2)$, see Figure \ref{fig14}a).

\section{Conclusion and outlook}\label{sec:5}

Significance of hyperbolic geometry for data science and ML is widely recognized in recent years. As experiments and novel approaches are actively investigated in various fields, there is an apparent necessity for systematic and adequate mathematical apparatus. This requires systematization of the previous mathematical knowledge, as well as development of new mathematical models and techniques.

We have presented some mathematical tools for learning in hyperbolic disc in Section \ref{sec:3} and applied them to RL problems in Section \ref{sec:4}. Our approach is based on explorations of data in hyperbolic spaces by learning their isometries. The same mathematical framework is relevant for supervised and unsupervised hyperbolic ML.

All problems considered here are solved by embedding the data into hyperbolic disc (i.e. in dimension two). The reasons for choosing such problems are transparency and easier visualization, rather than lack of models for dealing with problems in higher dimensions. In fact, models for  higher dimensions (most notably, for hyperbolic balls) are already available or can be elaborated. For instance, one can design and explore spatial directional labyrinths where an agent (or agents) learn directions in vector spaces of dimension three, or higher. An appropriate statistical model for learning orientations in high-dimensional vector spaces is presented, for instance, in \cite{KMcC}.

Mathematical tools from Section \ref{sec:3} are convenient for implementation of natural gradient stochastic policies (due to the fact that swarms from Section \ref{sec:3} are hyperbolic gradient flows in the unit disc), including, for instance algorithms in the spirit of natural Actor-Critic \cite{PS}.

We conclude the present paper with the statement that all preconditions are met for the phase transition in ML, with large-scale implementations of geometry-informed architectures and models in various fields of science, technology and everyday life. This will also result in reductions of the dimensionality and improved transparency of deep learning algorithms. Due to exponential growth of volume in hyperbolic spaces even moderate dimensions can be sufficient for very large data sets and demanding ML tasks.


\begin{thebibliography}{19}

\bibitem{Kuramoto} Kuramoto Y 1975 Self-entrainment of a population of coupled nonlinear oscillators {\it H. Araki (Eds.) International Symposium on Mathematical Problems in Theoretical Physics} (Springer Berlin Heidelberg) pp 420--422

\bibitem{Jacimovic} Ja\'cimovi\'c V 2024 Kuramoto Oscillators and Swarms on Manifolds for Geometry Informed Machine Learning {\it arXiv preprint arXiv:2405.09453}

\bibitem{HO} Hansen N and Ostermeier A 2001 Completely derandomized self-adaptation in evolution strategies {\it Evolutionary Computation} {\bf 9} 2 159--195 (MIT Press)

\bibitem{JA} James S and Abbeel P 2022 Bingham policy parameterization for 3{D} rotations in reinforcement learning {\it arXiv preprint arXiv:2202.03957}

\bibitem{CSS} Costa S I R, Santos S A and Strapasson J E 2015 Fisher information distance: {A} geometrical reading {\it Discrete Applied Mathematics} {\bf 197} 59--69 (Elsevier)

\bibitem{SHCSS} Salimans T, Ho J, Chen X, Sidor S and Sutskever I 2017 Evolution strategies as a scalable alternative to reinforcement learning {\it arXiv preprint arXiv:1703.03864}

\bibitem{Mettes} Mettes P, Ghadimi Atigh M, Keller-Ressel M, Gu J and Yeung S 2024 Hyperbolic deep learning in computer vision: {A} survey {\it International Journal of Computer Vision} pp 1--25 (Springer)

\bibitem{BBCV} Bronstein M M, Bruna J, Cohen T and Velickovic P 2021 Geometric deep learning: {G}rids, groups, graphs, geodesics, and gauges {\it arXiv preprint arXiv:2104.13478}

\bibitem{CCBH} Cetin E, Chamberlain B, Bronstein M and Hunt J J 2022 Hyperbolic deep reinforcement learning {\it arXiv preprint arXiv:2210.01542}

\bibitem{JM} Ja\'cimovi\'c V and Markovi\'c M 2024 Conformally Natural Families of Probability Distributions on Hyperbolic Disc with a View on Geometric Deep Learning {\it arXiv preprint arXiv:2407.16733}

\bibitem{PG-P} Pewsey A and Garc{\'\i}a-Portugu{\'e}s E 2021 Recent advances in directional statistics {\it Test} {\bf 30} 1 1--58 (Springer)

\bibitem{MJ} Mardia K V and Jupp P E 2000 {\it Directional statistics} (Chichester: John Wiley \& Sons)

\bibitem{McCullagh} McCullagh P 1996 M{\"o}bius transformation and {C}auchy parameter estimation {\it The Annals of Statistics} {\bf 24} 2 787--808 (Institute of Mathematical Statistics)

\bibitem{AG} Alexandrov A and Gorsky A 2023 Information geometry and synchronization phase transition in the {K}uramoto model {\it Physical Review E} {\bf 107} 4 044211 (APS)

\bibitem{SSGR} Sala F, De Sa C, Gu A and R{\'e} C 2018 Representation tradeoffs for hyperbolic embeddings {\it International Conference on Machine Learning} pp 4460--4469

\bibitem{TBG} Tifrea A, B{\'e}cigneul G and Ganea O-E 2018 Poincar\'e glove: {H}yperbolic word embeddings {\it arXiv preprint arXiv:1810.06546}

\bibitem{GBH} Ganea O, B{\'e}cigneul G and Hofmann T 2018 Hyperbolic neural networks {\it Advances in Neural Information Processing Systems} {\bf 31}

\bibitem{NK1} Nickel M and Kiela D 2017 Poincar{\'e} embeddings for learning hierarchical representations {\it Advances in Neural Information Processing Systems} {\bf 30}

\bibitem{NK2} Nickel M and Kiela D 2018 Learning continuous hierarchies in the {L}orentz model of hyperbolic geometry {\it International Conference on Machine Learning} pp 3779--3788

\bibitem{CHWDDV} Chamberlain, Benjamin Paul, Hardwick, Stephen R, Wardrope, David R, Dzogang, Fabon, Daolio, Fabio and Vargas, Saúl. 2019. Scalable hyperbolic recommender systems. {\it arXiv preprint arXiv:1902.08648}.

\bibitem{Baker} Baker, Cole, Suárez-Méndez, Isabel, Smith, Grace, Marsh, Elisabeth B, Funke, Michael, Mosher, John C, Maestú, Fernando, Xu, Mengjia and Pantazis, Dimitrios. 2024. Hyperbolic graph embedding of MEG brain networks to study brain alterations in individuals with subjective cognitive decline. {\it IEEE Journal of Biomedical and Health Informatics}, IEEE.

\bibitem{BSS} Barjuan, Laia, Soriano, Jordi and Serrano, M Ángeles. 2024. Optimal navigability of weighted human brain connectomes in physical space. {\it NeuroImage}, 297, 120703. Elsevier.

\bibitem{Chami} Chami, Ines, Abu-El-Haija, Sami, Perozzi, Bryan, Ré, Christopher and Murphy, Kevin. 2022. Machine learning on graphs: A model and comprehensive taxonomy. {\it Journal of Machine Learning Research}, 23(89), 1--64.

\bibitem{PS} Peters, Jan and Schaal, Stefan. 2008. Natural actor-critic. {\it Neurocomputing}, 71(7-9), 1180--1190. Elsevier.

\bibitem{KMcC} Kato, Shogo and McCullagh, Peter. 2020. Some properties of a Cauchy family on the sphere derived from the Möbius transformations. {\it Bernoulli}, 26(4), 3224--3248. Bernoulli Society for Mathematical Statistics and Probability.

\bibitem{Needham} Needham, Tristan. 2023. Visual complex analysis. Oxford: Oxford University Press.

\bibitem{AJLS} Ay, Nihat, Jost, J\"{u}rgen, V\^{a}n L\^{e}, H\`{o}ng and Schwachh\"{o}fer, Lorenz. 2017. Information geometry. 64. Berlin: Springer.

\bibitem{MMS} Marvel, Seth A, Mirollo, Renato E and Strogatz, Steven H. 2009. Identical phase oscillators with global sinusoidal coupling evolve by Möbius group action. {\it Chaos: An Interdisciplinary Journal of Nonlinear Science}, 19(4). AIP Publishing.

\bibitem{BAORMK} Bogatskiy, Alexander, Anderson, Brandon, Offermann, Jan, Roussi, Marwah, Miller, David and Kondor, Risi. 2020. Lorentz group equivariant neural network for particle physics. In: {\it International Conference on Machine Learning}, pp. 992--1002.

\bibitem{KPKVB} Krioukov, Dmitri, Papadopoulos, Fragkiskos, Kitsak, Maksim, Vahdat, Amin and Boguná, Marián. 2010. Hyperbolic geometry of complex networks. {\it Physical Review E}, 82(3), 036106. APS.

\bibitem{BFKL} Bläsius, Thomas, Friedrich, Tobias, Krohmer, Anton and Laue, Sören. 2018. Efficient embedding of scale-free graphs in the hyperbolic plane. {\it IEEE/ACM Transactions on Networking}, 26(2), 920--933. IEEE.
\end{thebibliography}





\end{document}